\documentclass{article}

    \PassOptionsToPackage{numbers, compress}{natbib}

\usepackage[preprint]{neurips_2026}

\usepackage[utf8]{inputenc} 
\usepackage[T1]{fontenc}    
\usepackage{hyperref}       
\usepackage{url}            
\usepackage{booktabs}       
\usepackage{amsfonts}       
\usepackage{nicefrac}       
\usepackage{microtype}      
\usepackage{xcolor}         
\usepackage{graphicx}
\usepackage{amsmath}
\usepackage{xspace}
\usepackage{enumitem}  
\usepackage{multirow}
\usepackage{wasysym} 
\usepackage{pifont}
\usepackage{tabularray}
\usepackage[table]{xcolor}
\usepackage[most]{tcolorbox}
\usepackage{fvextra}
\fvset{breaklines=true,breakanywhere=true} 
\usepackage{wrapfig}
\usepackage{rotating}
\usepackage{etoc}

\title{Cookie-Bench: Continuous On-screen Key Interaction Evaluation for Web Generation}

\author{%
  \bfseries Haoyue Yang\thanks{Equal contribution.}, Zhangxiao Shen\footnotemark[1], Fan Ding\footnotemark[1], Hangting Lou, Yifeng Kou, \\[3pt]
  \bfseries Haoqing Yu, Jingyao Li\thanks{Corresponding author}, Zhengfan Wu\footnotemark[2], Siqi Bao\footnotemark[2], Jing Liu, Hua Wu\\[6pt]
  \normalfont Baidu Inc., Beijing, China\\[6pt]
  \small\texttt{\{yanghaoyue, shenzhangxiao\}@baidu.com}
}

\begin{document}

\newcommand{\papername}{Cookie-Bench\xspace}
\newcommand{\framename}{Cookie\xspace}   
\newcommand{\dataname}{Cookie-Bench\xspace} 
\newcommand{\benchname}{Cookie-Bench\xspace}

\maketitle

\vspace{-3mm}
\begin{abstract}
\vspace{-3mm}
Front-end web code has become a core product surface for every frontier LLM release, yet evaluating these interactive applications at development speed remains costly because human-judged leaderboards like Arena do not scale.  Existing automated proxies typically lean on reference implementations, test suites, or rigid checklists, and tend to miss the reasoned synthesis a human reviewer performs over a live session. We articulate a new evaluation regime that is simultaneously reference-free, autonomously driven, and holistically reasoned, and instantiate it through two artifacts. \textbf{\dataname} is an 11-domain, 54-leaf, 1,000-query WebDev benchmark spanning both static-presentation and interactive-application tasks, balanced across three difficulty tiers and three target-language groups, with briefs rewritten to resist recall from circulated prompts. \textbf{\framename}, grounded in Flavell's metacognitive monitoring, separates evidence accumulation from judgment across three stages: Static Perception forms a first impression from passive observation; Agent-Driven Interaction explores the application autonomously while capturing continuous screen video, audio, and per-step screenshots; Dynamic Scoring issues holistic functionality and aesthetics verdicts with structured failure attribution only after the evidence chain is complete. On \dataname, \framename aligns closely with expert human ratings while surfacing substantial headroom across 13 frontier LLMs on interactive web generation. \noindent\url{https://github.com/Haoyue-Yang/Cookie}
\end{abstract}

\begin{figure}[htbp]
  \centering
  \vspace{-2mm}
  \includegraphics[width=0.83\linewidth]{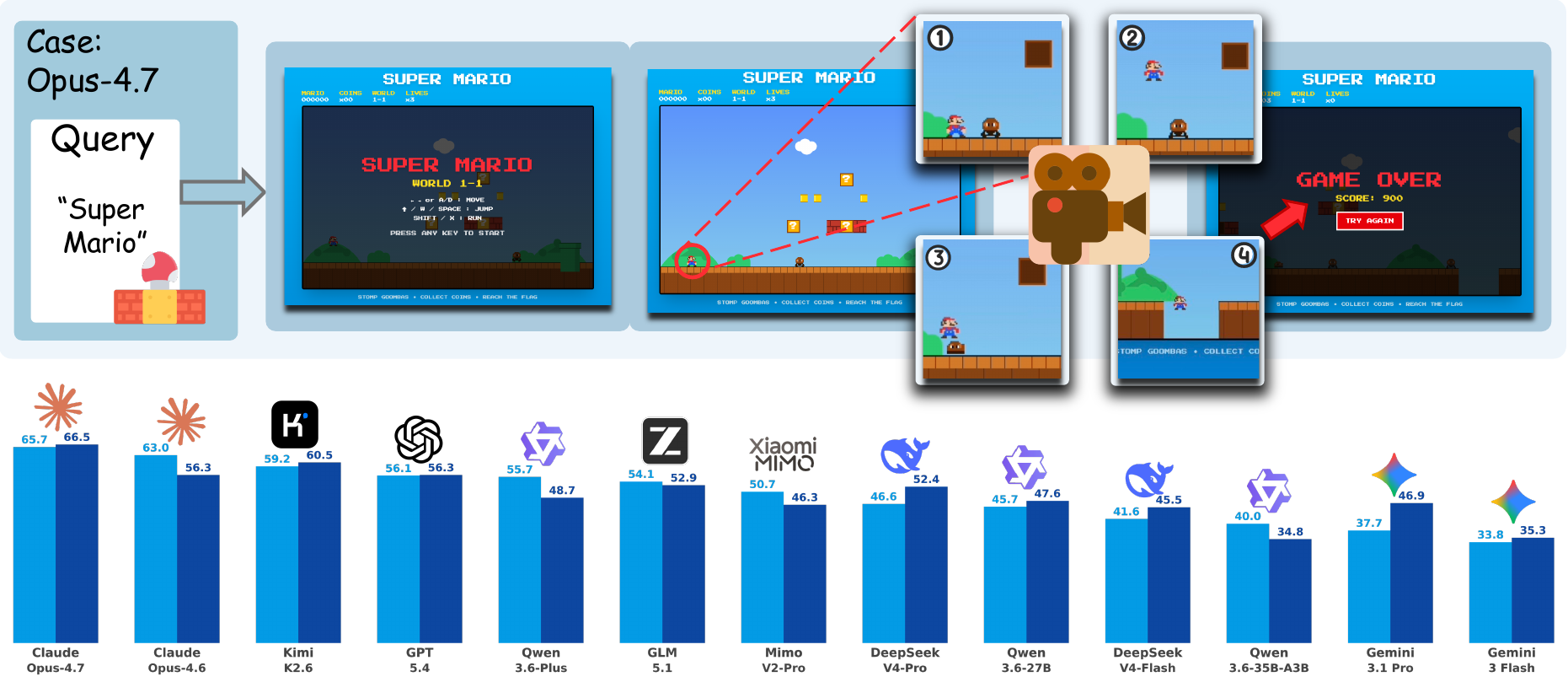}
  \caption{
    Top: Query ``Super Mario'' flowing through 
    deployment, autonomous agent-driven interaction, and multi-modal evidence capture. Bottom: overall total Win rate for 13 frontier LLMs on \dataname; left bars: agent-scaffolded React generation; right bars: direct HTML chat output.}
  \vspace{-2mm}
  \label{fig:showcase_rank}
\end{figure}

\section{Introduction}
\label{sec:intro}
\vspace{-2mm}

Recent months have seen frontier LLMs from OpenAI~\citep{openai2026gpt54}, Anthropic~\citep{anthropic2026claudevisuals}, and Google~\citep{google2026gemini3} race to ship interactive web-rendered responses as a core product capability. This arms race has quietly made the evaluation of the scarce resource. Arena-style leaderboards—Chatbot Arena, Code Arena, and, more recently, WebDev Arena and Design Arena—remain the gold standard for judging user-perceivable quality because they route verdicts through actual humans. But a single Arena ranking round is prohibitively expensive in paid labor and takes weeks to settle, making Arena unusable as an inner-loop signal during model development. An affordable automated proxy that preserves Arena-level judgment fidelity is therefore an urgent need. Existing automated web-code benchmarks fall short of this bar. Whether they score a single rendered frame~\citep{laurenccon2024unlocking, si2025design2code, gui2025webcode2m, sun2025fullfront, xiao2025designbench, liu2026webcoderbench} or exercise the page along a fixed~\citep{xu2025web, lu2025webgen, zhang2026miniappbench} or agent-planned~\citep{xiao2025code, lei2026webcompassmultimodalwebcoding} trajectory, they all lean on some combination of reference implementations, pre-authored test suites, and rigid checklist aggregation, and none reproduces the reasoned synthesis a human reviewer performs over a live session.

We target the missing combination directly. A competent human reviewer, in the spirit of Flavell's metacognitive monitoring~\citep{flavell1979metacognition}, draws on internalized \emph{knowledge} of what a page should look like, accumulates real-time \emph{experience} by interacting with it, and \emph{regulates} judgment by reasoning over that evidence—without consulting any reference. Our evaluator, \textbf{\framename}, instantiates this loop as three sequential stages. \emph{Static Perception} loads the deployed page and forms a first impression from passive observation. \emph{Agent-Driven Interaction} hands control to a computer-using agent that autonomously plans an exploration trajectory and records screenshots, a continuous screen recording, an audio track, and an interaction trace, adapting on the fly to pursue unexpected paths. \emph{Dynamic Scoring} defers all evaluative reasoning until the full evidence package is assembled, then synthesizes holistic functionality and aesthetics scores with structured failure attribution. No reference implementation, test suite, or pre-authored checklist is consulted at any stage. Figure~\ref{fig:showcase_rank} top illustrates this pipeline on a representative ``Super Mario'' query. The static stage confirms that all mechanics are implemented and awards a high score, but it cannot simulate the physics of the interaction; the interaction video shows that the jump distance is too short for the gap.

Paired with \framename is \textbf{\dataname}, an 11-domain, 1,000-query WebDev benchmark. Queries span both static-presentation and interactive-application tasks, are balanced across three difficulty tiers and three target-language groups, and are rewritten into self-contained, engineering-feasible briefs that resist recall from publicly circulated prompts, so that passing \dataname reflects capability rather than memorization. Table~\ref{tab:benchmark_comparison} positions \papername against representative benchmarks from each regime above: to our knowledge, it is the only setup that simultaneously covers both static and dynamic tasks, drives evaluation autonomously, verifies behavior on continuous execution rather than discrete frames, and does so without external reference of any kind. Figure~\ref{fig:showcase_rank} bottom reports the resulting Total scores for 13 frontier LLMs, with the React-to-HTML gap itself serving as a model-dependent signal of architectural competence.

Our contributions are as follows. We articulate a new evaluation regime for web code that is simultaneously \emph{reference-free}, \emph{autonomously driven}, and \emph{holistically reasoned}, shedding reference implementations, test scripts, and pre-authored checklists at once rather than one at a time. We instantiate this regime through two artifacts: \textbf{\dataname}, described above, and \textbf{\framename}, a three-stage agent-driven evaluator that separates evidence accumulation from judgment and synthesizes multi-modal evidence into holistic scores with failure attribution. Using \framename, we evaluate 13 frontier LLMs on \dataname along functionality and aesthetics across three difficulty tiers. 

\section{Related Work}

\begin{table}[htbp]
\centering
\caption{Representative text-to-webcode benchmarks along the axes introduced in this section.}
\label{tab:benchmark_comparison}
\resizebox{\columnwidth}{!}{
\begin{tabular}{l|ccccccc}
\toprule
\textbf{Benchmark} & \textbf{Task} & \textbf{Driving} & \textbf{Verify} & \textbf{Ref.} & \textbf{Mode} & \textbf{\# Instances} \\
\midrule
WebCoderBench~\cite{liu2026webcoderbench}       & Static          & Ref-Guided          & Screenshot & Checklist     & Agentic\&Chat & 1572 \\
Web-Bench~\cite{xu2025web}           & Both         & Pre-defined  & E2E Test  & Ground truth  & Agentic     & 50\ \\
WebGen-Bench~\cite{lu2025webgen}        & Both         & Ref-Guided   & Multi-frame  & Checklist     & Agentic     & 101     \\
WebCompass~\cite{lei2026webcompassmultimodalwebcoding}          & Both          & Autonomous  & Multi-frame  & Checklist     & Chat & 123 \\
OpenDesignBench~\cite{xiao2025code}     & Both         & Autonomous   & Multi-frame  & w/o Ref.      & Chat           & 840     \\
\rowcolor[HTML]{D8ECE3} COOIKE-Bench & Both & Autonomous   & Video   & w/o Ref.      & Agentic\&Chat & 1000     \\
\bottomrule
\end{tabular}
\vspace{-5mm}
}
\end{table}

\paragraph{Static Evaluation.}
Static-evaluation benchmarks render the generated page once and score it on a single frame. Early work uses deterministic similarity metrics: WebSight~\citep{laurenccon2024unlocking} and Web2Code~\citep{yun2024web2code} report pixel- and text-level scores (SSIM, BLEU); Design2Code~\citep{si2025design2code} adds CLIP embeddings; and WebCode2M~\citep{gui2025webcode2m}, Vision2UI~\citep{gui2024vision2ui}, and IW-Bench~\citep{guo2025iw} extend the comparison to element-level granularity. Subsequent work replaces these metrics with LLM- or VLM-based judges under three verification regimes: reference-grounded scoring against a ground-truth design in FullFront~\citep{sun2025fullfront}, DesignBench~\citep{xiao2025designbench}, and WebRenderBench~\citep{lai2025webrenderbench}; reference-free query-image similarity in UIClip~\citep{wu2024uiclip}; and rubric-based scoring against human-authored checklists in UI-Bench~\citep{jung2025ui}, which calibrates rubric judges against human experts, and WebCoderBench~\citep{liu2026webcoderbench}, which combines rule-based analyzers (linters, Lighthouse audits, static syntax/accessibility checks) with LLM-as-a-judge scoring over 24 fine-grained metrics on the rendered code and screenshot. In all cases, evaluation observes a single rendered frame and does not interact with the page; interactive components and runtime state transitions are therefore outside the evaluation scope.

\vspace{-3mm}
\paragraph{Dynamic Evaluation with Pre-Defined Driving.}
Dynamic benchmarks in this group run the page in a browser and drive it along a trajectory fixed in advance, either as executable code or as a per-query reference. Interaction2Code~\citep{xiao2025interaction2code} compares pre/post-interaction screenshots under fixed actions; ArtifactsBench~\citep{zhang2025artifactsbench} captures three staged screenshots around a scripted interaction and scores them with an MLLM referee under a fine-grained checklist; Web-Bench~\citep{xu2025web}, FrontendBench~\citep{zhu2025frontendbench}, MRWeb~\citep{wan2024mrweb}, and IWR-Bench~\citep{chen2025iwr} rely on end-to-end test suites (Playwright, Jest, Selenium) targeting DOM, visual, or behavioral properties. A second wave delegates action execution to an agent but retains a human-authored target: WebGen-Bench~\citep{lu2025webgen}, AUI-Gym~\citep{qinghong2025computer}, WebVIA~\citep{xu2025webvia}, and FullStack-Bench~\citep{lu2026fullstack} pair navigation or Computer-Use agents with pre-specified checks. Vision2Web~\citep{he2026vision2web} and MiniAppBench~\citep{zhang2026miniappbench} push this further with structured per-query references: Vision2Web constrains a WebVoyager-based GUI agent with an expert-designed workflow graph whose nodes are 3-tuples $\langle O_i, A_i, V_i \rangle$ of objective, guided actions, and validation criteria, and MiniAppBench pairs each query with a reference $r_i$ enumerating verifiable points across intention, static, and dynamic dimensions. Across this group, the verification scope is bounded by the human-specified script or reference.

\vspace{-3mm}
\paragraph{Dynamic Evaluation with Autonomous Driving.}
Here, the agent generates the trajectory from the query alone. WebTestBench~\citep{kong2026webtestbench} targets testing of pre-built apps, where the agent decomposes the instruction into a checklist and reaches a Pass/Fail verdict per item; OpenDesign Bench~\citep{xiao2025code}, released alongside a training recipe for an aesthetics-focused code model rather than as a standalone evaluation suite, uses a WebVoyager-based agent that ranks and executes interaction candidates for aesthetic scoring; WebCompass~\citep{lei2026webcompassmultimodalwebcoding} adopts an Agent-as-a-Judge pipeline that pairs an LLM-generated checklist with Claude Code and an MCP-controlled browser for interaction and adaptive JavaScript test synthesis. Beyond the web, PlayCoder~\citep{peng2026playcodermakingllmgeneratedgui} applies this protocol to a desktop GUI within a broader GUI-code training system, verifying via final-state frames and logs under Play@k.


Table~\ref{tab:benchmark_comparison} summarizes representative benchmarks along these axes: task type, evaluation protocol, generation mode, and scale. The evaluation protocol spans how the page is driven, what evidence is verified, and whether a reference is required. \dataname{} covers both static and dynamic generation tasks, supports both chat-direct and scaffolded code production, generates the interaction trajectory autonomously, and verifies behavior on continuous execution video under human-calibrated rubrics, without relying on a ground-truth implementation or a per-query checklist.

\vspace{-2mm}
\section{\dataname Benchmark Data}
\label{sec:data}
\vspace{-2mm}

\begin{figure}[htbp]
  \centering
  \vspace{-3mm}
  \includegraphics[width=\linewidth]{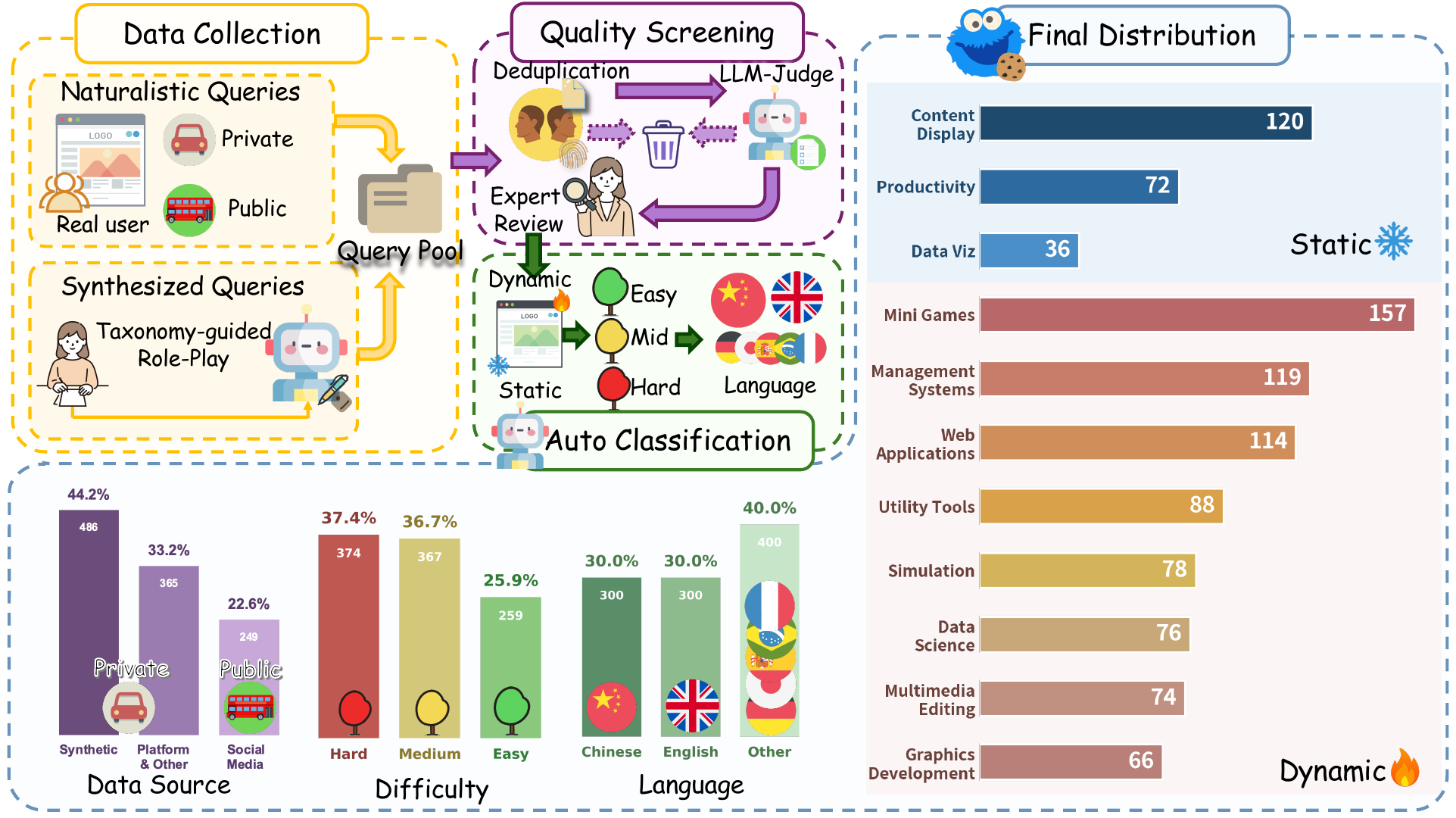}
  \caption{
    \textbf{\dataname data construction pipeline and dataset statistics.} The upper-left shows the data construction pipeline; the lower-right shows the dataset distribution statistics.}
    \vspace{-4mm}
  \label{fig:data_stats_new}
\end{figure}

Constructing a WebDev benchmark forces three design questions in sequence: which scenarios should the benchmark cover, how should the admitted samples be distributed so that capability gaps are interpretable, and how can every sample be labeled consistently at leaf granularity. The subsections below address these questions in turn; full protocols, tool implementations, and validation numbers are consolidated in Appendix~\ref{app:data}.

\vspace{-4mm}
\subsection{Data Sources and Quality Assurance}
\label{sec:data_sources}
\vspace{-2mm}

A WebDev benchmark must cover every sub-skill a practicing front-end engineer exercises, and probe the capability boundary of current models rather than their recall of already-circulated prompts. Inheriting the raw pool's category structure undercovers tail scenarios, and relying solely on open-web queries conflates memorization with capability. \dataname therefore fixes an 11-domain, 54-leaf WebDev taxonomy up front as the scenario scaffold, and populates each leaf from two complementary sourcing regimes reported in Figure~\ref{fig:data_stats_new}.

No single sourcing channel satisfies both requirements alone, so the two regimes are chosen to cover each other's blind spots. \emph{Naturalistic queries}, contributing 514 entries drawn from real-user traffic on an internal WebDev product and public user-evaluation channels, pre-exist the benchmark and preserve authentic user intent, colloquial phrasing, and incomplete specifications that synthetic authoring cannot fabricate. \emph{Crowd-synthesized queries}, contributing the remaining 486 entries and labeled \emph{Synthetic} in the figure, are authored at benchmark construction time under a taxonomy-guided role-play protocol, yielding uniform coverage of tail leaves and prompts unlikely to appear in any pretraining corpus. Raw queries from both channels then pass through a three-stage quality pipeline: a two-layer deduplication combining a SimHash-based~\citep{sadowski2007simhash} lexical pass and a TF-IDF-based~\citep{robertson2004understanding} semantic pass, with every merge logged for sample-level audit; an LLM-judge filter along seven independently checkable admissibility axes covering safety, privacy leakage, front-end scope, executability, external-dependency minimality, intent clarity, and logical completeness; and an expert-review stage reserved for judgment-heavy decisions on difficulty, taxonomy placement, and borderline admissibility, together with an audit of samples from the upstream stages. The admitted pool contains 1,000 queries, to which all statistics below refer.

\vspace{-3mm}
\subsection{Data Composition and Statistics}
\label{sec:data_composition}
\vspace{-2mm}

Once admitted, the 1,000 queries must be distributed so that capability gaps along any one axis can be measured while the other axes are held approximately constant. This design objective, rather than mere reporting convenience, dictates the distributions shown in Figure~\ref{fig:data_stats_new}; we motivate the three axes in turn.

Static and dynamic pages demand categorically different capabilities: static pages exercise layout, typography, and visual composition at render time, whereas dynamic pages exercise state management, event handling, and cross-component coordination over user interaction. We split at L1 and group the 11 L2 domains by shared output artifact, with static covering presentation-oriented surfaces such as content display, data reporting, and marketing pages, and dynamic covering interaction-driven surfaces such as tools, dashboards, games, and simulations. A flat easy/medium/hard tag is unreliable, so we score every query on six orthogonal dimensions covering functional logic, page and interaction complexity, data and system demands, visual design, user experience, and dynamic simulation, and take the maximum as the overall difficulty. The resulting split concentrates on Medium and High with few trivial prompts, matching our goal of probing the frontier of model capability rather than saturating on easy cases. Real-user WebDev queries arrive in many languages, but the raw pool is dominated by Chinese and English. We rebalance to an average split across Chinese, English, and six additional widely used languages (French, Spanish, Japanese, German, Korean, Portuguese), using stratified retention on the $\text{L1}\times\text{L2}\times\text{difficulty}$ key so that the taxonomy and difficulty distributions on each language remain identical.

\vspace{-2mm}
\subsection{Automated Data Classification Pipeline}
\label{sec:pipeline}
\vspace{-2mm}

The taxonomy, difficulty, and language distributions of Section~\ref{sec:data_composition} all rely on one operation: assigning each query to its L3 leaf. Manual labeling across 54 semantically adjacent leaves drifts with panel fatigue, so we split the work by comparative advantage, pushing leaf selection onto an LLM under low-temperature decoding and reserving humans for definition design and borderline arbitration.

The pipeline maps each query to one leaf in a single low-temperature forward pass. The prompt inlines the full 54-leaf definition list and instructs the model to classify on the underlying task scenario rather than surface wording, with L1 and L2 recovered deterministically from the taxonomy tree. Because a single pass conflates genuine model errors with errors induced by underspecified leaf definitions, we wrap this pass in a feedback loop: run the pipeline on a human-labeled validation slice, cluster residual errors by gold--predicted leaf pair, sharpen the confused definitions, and re-run on the full benchmark. Expert review spot-checks the revised labels before admission. Prompt template, validation protocol, error tables, and revised definitions are in Appendix~\ref{app:data}.

\vspace{-3mm}
\section{Evaluation Methodology}
\label{sec:eval_method}
\vspace{-2mm}

Human reviewers judge websites via \emph{metacognitive monitoring}~\citep{flavell1979metacognition}: they apply internalized quality priors to evidence from observation and interaction, then regulate verdicts through reasoning. \dataname instantiates this as \framename, whose judge uses the same global, application-agnostic scoring prior across all queries. The subsections below describe the framework and its two dimensions; full protocols are in Appendix~\ref{app:method}.

\vspace{-3mm}
\subsection{Agent-Driven Interactive Evaluation}
\label{sec:eval_framework}
\vspace{-2mm}

\begin{figure}[t]
  \centering
  \includegraphics[width=\columnwidth]{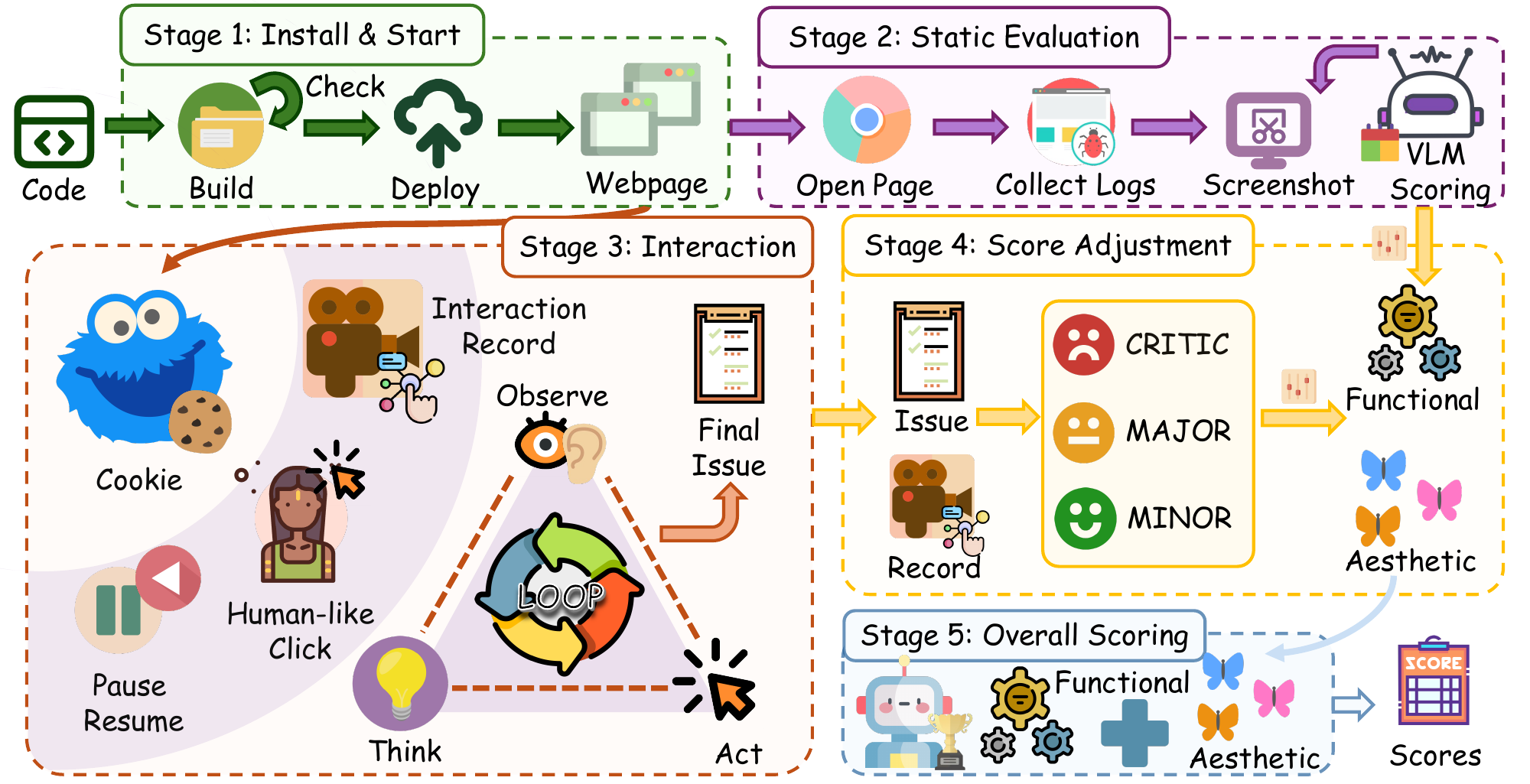}
  \vspace{-2mm}
  \caption{\textbf{Overview of \framename.} A five-stage pipeline from code to score: \emph{Install \& Start} deploys the generated page; \emph{Static Evaluation} captures logs and a VLM-scored screenshot; \emph{Interaction} runs the Cookie agent through an Observe-Think-Act loop with human-like clicks; \emph{Score Adjustment} grades issues at Critical, Major, or Minor severity across Functional and Aesthetic dimensions; \emph{Overall Scoring} aggregates them into the final score.}
  \label{fig:eval_pipeline}
  \vspace{-2mm}
\end{figure}

Given a user query $q_i$, the first two stages instantiate \emph{metacognitive experience} by progressively accumulating multi-modal evidence, while the third instantiates \emph{metacognitive regulation} through holistic reasoning over that evidence.

A first impression is informative but easy to overtrust, and much of what looks right at a glance breaks the moment a user interacts with it. \framename therefore opens with \emph{Static Perception}, which loads the deployed application and collects evidence without user action, i.e., a full-page screenshot, runtime error traces, and a structural inventory of the rendered interface, then passes it to a vision-language model that emits provisional per-dimension scores. These scores are retained as a prior belief, subject to revision once interactive evidence arrives.

Many defects of a web artifact, including broken forms, stale state, and hidden flows, only manifest under input and remain invisible to any amount of passive observation. \emph{Agent-Driven Interaction} addresses these through \framename, our computer-using evaluation agent, which autonomously plans an exploration trajectory given $q_i$ and executes it via an \emph{observation--thought--action} loop that adapts on the fly to pursue unexpected paths and revisit suspicious behaviors. Throughout the session, a multi-modal capture pipeline records screen video, audio, and per-step screenshots, preserving temporal dynamics that static snapshots lose. The stage yields an evidence package of trajectory, keyframes, continuous recording, and an agent-generated problem summary.

Scoring a stage at a time invites confirmation bias, since any judgment emitted mid-interaction anchors every later observation toward it~\citep{flavell1979metacognition}. \emph{Dynamic Scoring} therefore defers all evaluative reasoning until the evidence chain is complete, at which point it analyzes the full package, surfaces defects not apparent from passive observation, and uses them to confirm or overturn the Stage~1 prior. The output is a pair of calibrated per-dimension scores with structured failure attribution identifying what failed, where it manifested, and why it constitutes a defect. The two dimensions are defined in the next subsection.

\vspace{-2mm}
\subsection{Evaluation Dimensions}
\label{sec:eval_dims}
\vspace{-2mm}

The final output of \framename is a judge agent that, after the \emph{Dynamic Scoring} stage, emits one scalar per web artifact per dimension. Two design choices determine this score: which dimensions the agent reports on, and what form each per-dimension score takes.

For the first choice, prior web-code benchmarks span granularities from a single pass rate~\citep{xu2025web} to twenty-four metrics under nine perspectives~\citep{liu2026webcoderbench}, and we adopt two desiderata: an axis should be \emph{high-level enough} to remain stable across diverse web scenes, and \emph{orthogonal enough} that a given failure falls cleanly on one side rather than spreading across several. Under these desiderata, we report \textbf{functionality} and \textbf{aesthetics}, and file interactivity inside this pair: interactive-logic correctness is a semantic behavior we score under functionality, while transition naturalness is a perceptual behavior we score under aesthetics.

For the second choice, we let the judge agent emit a single holistic continuous score per dimension rather than an averaged checklist of sub-criteria. The reasoning is that a vision-language judge consumes a composite evidence package, including screenshots, interaction trajectory, runtime traces, and agent-generated problem summaries, and its strength lies in weighing the severity of failures against one another across this body rather than in ticking individual boxes; a rigid checklist would collapse that severity weighing into a sum of equal-weight items and discard the holistic reasoning that makes a strong reviewer valuable.

To cross-check the machine judge, we additionally collect human ratings on a small held-out slice using a sixteen-item binary rubric, deliberately finer-grained and designed for inter-rater consistency. The rubric, together with the judge agent's scoring prompts, its scene-adapted criteria, exemption rules, and aggregation formula, is deferred to Appendix~\ref{app:method}; both the rubric and the prompts are fixed before any query is drawn and shared across the benchmark, rather than authored per item against a reference.

\vspace{-2mm}
\section{Experiments}
\label{sec:expe}
\vspace{-2mm}

We evaluate 13 frontier LLMs on \dataname, spanning Claude~\cite{claude-opus-4-7,claude-opus-4-6}, GPT~\cite{gpt-5-4}, Kimi~\cite{kimi-k2-6}, GLM~\cite{glm-5-1}, Qwen, Mimo~\cite{mimo-v2-pro}, DeepSeek~\cite{deepseek-v4}, and Gemini~\cite{gemini-3-1}. The selection ranges from 27B-parameter open-weight models to closed-source APIs, with all models invoked through standard APIs using identical system prompts and the same queries drawn from \dataname.

Each model is assessed under two generation settings. In the \textbf{React} setting, the model operates within an agent scaffold and modifies it through tool calls to implement the query. In the \textbf{HTML} setting, the model receives only the user query and outputs a single self-contained HTML file without any scaffold or build pipeline. The scaffold exposes file operations (create, read, edit, delete, list, glob, grep, patch), project execution (build, npm install), web access (search, fetch), image generation, and plan updates; HTML mode relies solely on in-context generation without external tool access. Full scaffold and tool specifications are in Appendix~\ref{app:scaffold}. Both outputs are evaluated by \framename, which synthesizes holistic functionality and aesthetics scores from static perception, agent-driven interaction, and dynamic scoring.

\begin{table*}[t]
\centering
\small
\caption{Main results across two generation modes (React agent scaffold vs.~HTML direct chat), two page types (Dynamic, Static), and three evaluation dimensions (Functionality, Aesthetics, Total). \textbf{Bold} denotes the best score per column; \underline{underline} denotes the second best.}
\label{tab:main_dynamic_static}
\setlength{\tabcolsep}{1pt}
\resizebox{\textwidth}{!}{
\begin{tabular}{lcccccccccccccc}
\toprule
\multirow{3}{*}{\textbf{Model}}
  & \multicolumn{7}{c}{\textbf{React (agent scaffold)}}
  & \multicolumn{7}{c}{\textbf{HTML (direct chat)}} \\
\cmidrule(lr){2-8}\cmidrule(lr){9-15}
  & \multicolumn{3}{c}{\textbf{Dynamic}}
  & \multicolumn{3}{c}{\textbf{Static}}
  & \textbf{Overall}
  & \multicolumn{3}{c}{\textbf{Dynamic}}
  & \multicolumn{3}{c}{\textbf{Static}}
  & \textbf{Overall} \\
\cmidrule(lr){2-4}\cmidrule(lr){5-7}\cmidrule(lr){9-11}\cmidrule(lr){12-14}
  & Func. & Aes. & Total & Func. & Aes. & Total &  & Func. & Aes. & Total & Func. & Aes. & Total &  \\
\midrule
Qwen3.6-35B-A3B      & 67.3 & 70.1 & 59.8 & 73.2 & 74.6 & 66.2 & 61.3 & 67.7 & 80.8 & 61.7 & 74.6 & 84.4 & 67.7 & 63.1 \\
Gemini-3-Flash       & 68.8 & 82.8 & 63.5 & 70.3 & 79.6 & 62.8 & 63.3 & 74.3 & 78.8 & 65.5 & 78.1 & 78.7 & 68.2 & 66.1 \\
Gemini-3.1-Pro       & 70.8 & 80.3 & 64.3 & 75.5 & 79.0 & 66.7 & 64.9 & 80.5 & 85.7 & 73.7 & 80.9 & 83.4 & 72.8 & 73.5 \\
Qwen3.6-27B          & 75.6 & 74.2 & 66.3 & 76.7 & 78.5 & 69.9 & 67.2 & 73.3 & 87.2 & 69.6 & 78.6 & 89.9 & 74.3 & 70.7 \\
DS-V4-Flash          & 71.1 & 79.5 & 66.2 & 78.6 & 87.8 & 73.6 & 67.9 & 78.1 & 87.1 & 72.4 & 81.4 & 83.7 & 73.3 & 72.6 \\
DS-V4-Pro            & 76.8 & 87.7 & 72.1 & 81.2 & 89.2 & 76.4 & 73.2 & 80.9 & 90.4 & 76.8 & 71.9 & 79.1 & 67.9 & 74.6 \\
Qwen3.6-Plus         & 83.9 & 79.9 & 74.8 & 83.4 & 83.6 & 76.6 & 75.2 & 78.7 & 90.3 & 74.3 & 79.8 & 89.4 & 74.7 & 74.4 \\
GLM-5.1              & 79.5 & 85.7 & 74.6 & 86.4 & 90.1 & 81.5 & 76.2 & 80.6 & 90.5 & 76.5 & 80.4 & 87.1 & 75.0 & 76.1 \\
Mimo-V2-Pro          & 82.1 & 84.7 & 75.2 & 85.9 & 88.7 & 80.2 & 76.3 & 79.9 & 87.5 & 73.8 & 81.0 & 86.4 & 74.3 & 73.9 \\
GPT-5.4              & 80.5 & 90.5 & 76.4 & 87.1 & \underline{91.4} & \underline{82.6} & 77.8 & 80.9 & 90.0 & 76.9 & \underline{88.1} & \underline{92.3} & \underline{84.1} & 78.6 \\
Kimi-K2.6            & 81.6 & \underline{91.1} & 78.9 & 81.3 & 87.1 & 76.6 & 78.3 & \underline{85.0} & 88.6 & \underline{80.8} & 78.9 & 85.2 & 72.7 & \underline{78.9} \\
Opus 4.6             & \textbf{87.8} & 84.7 & \underline{79.4} & \textbf{87.5} & 88.7 & 82.2 & \underline{80.0} & 82.2 & \textbf{93.2} & 79.0 & 82.7 & 91.5 & 78.5 & \underline{78.9} \\
Opus 4.7             & \underline{85.8} & \textbf{94.1} & \textbf{82.9} & \underline{87.3} & \textbf{94.8} & \textbf{84.7} & \textbf{83.3} & \textbf{87.8} & \underline{91.4} & \textbf{83.5} & \textbf{90.9} & \textbf{92.3} & \textbf{86.7} & \textbf{84.2} \\
\bottomrule
\end{tabular}
}
\vspace{-2mm}
\end{table*}

\vspace{-2mm}
\subsection{Main Results}
\label{sec:main_results}

Table~\ref{tab:main_dynamic_static} reports \framename scores. Claude-Opus-4.7 leads React (83.3) and HTML (84.2), followed by Claude-Opus-4.6 (80.0 / 78.9), Kimi-K2.6(78.3 / 78.9) and GPT-5.4 (77.8 / 78.6). The 22-point React spread and 21-point HTML spread show scaffold complexity raises the floor for weaker models. Runnable rates exceed 94.9\% in React, confirming scaffold stability.

\textbf{Static versus Dynamic.} The scaffold reshapes which task types models handle well. In React, 11 of 13 models score higher on static pages, with DeepSeek-V4-Flash (+7.4), GLM-5.1 (+6.9), and Mimo-V2-Pro (+5.0) showing the largest gaps, because stateful interaction under a component architecture demands event wiring and lifecycle management that weaker models struggle to coordinate. Kimi-K2.6 is the sole exception (dynamic +2.3). In HTML the pattern inverts: seven models score higher on dynamic pages, led by DeepSeek-V4-Pro (+8.9) and Kimi-K2.6 (+8.1), because raw HTML/JS expresses stateful interaction more directly than React component decomposition. GPT-5.4 is the only major exception ($-$7.2), indicating its strength lies in layout rather than interaction logic.

\begin{figure*}[t]
\centering
\vspace{-2mm}
\includegraphics[width=\textwidth]{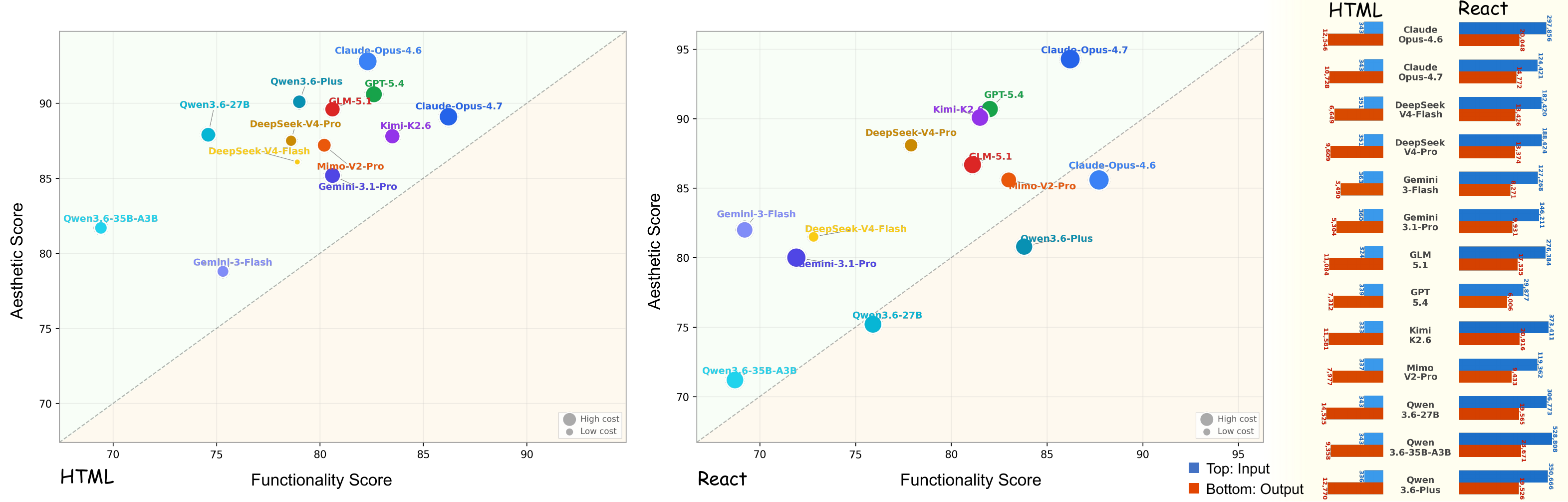}
\caption{Model capability landscape on \dataname. Left: Per-model Functionality--Aesthetics scatter plots, HTML on top and React on bottom; marker area encodes estimated per-query API cost. Right: Per-model input (blue) and output (orange) token counts, HTML on top and React on bottom.}
\label{fig:main-results}
\vspace{-4mm}
\end{figure*}

\textbf{Capability landscape and model preferences.} Figure~\ref{fig:main-results} left maps each model on the Functionality--Aesthetics plane under both modes, with marker area encoding per-query API cost. Under React the cloud elongates along the aesthetics axis: Claude-Opus-4.7 trades functionality for aesthetics versus Claude-Opus-4.6. Under HTML the entire cloud shifts above the identity line, confirming a visual-priority bias; Claude-Opus-4.6 reaches 92.8 aesthetics, the highest score across both modes. The React-to-HTML shift is family-dependent. Claude-Opus-4.7 (+0.9) and GPT-5.4 (+0.8) stay near-zero, showing architectural competence that does not relax in simpler settings. Claude-Opus-4.6 declines ($-$1.1), as its functional precision is underutilized without component decomposition. Gemini-3.1-Pro (+8.6), DeepSeek-V4-Flash (+4.7), Qwen3.6-27B (+3.5), and Gemini-3-Flash (+2.8) all rise in HTML, while Mimo-V2-Pro ($-$2.4) and Qwen3.6-Plus ($-$0.8) decline. Most mid-tier models improve in HTML; the scaffold's benefit is model-dependent, not universal. The cost overlay reveals steep diminishing returns: Claude-Opus-4.6 in React costs \$802.97 in total, roughly 120$\times$ DeepSeek-V4-Flash at \$6.75, yet the score advantage over mid-tier models stays within single digits.

\textbf{Token and cost structure.} Figure~\ref{fig:main-results} right quantifies the scaffold penalty. HTML demands only $\sim$340 input tokens per query, whereas React scaffolding consumes 30K--530K input tokens. Output tokens are more comparable, $\sim$7K--24K in React versus $\sim$3.5K--14.5K in HTML. This input asymmetry translates directly into cost: React totals 2.5--3.5$\times$ more than HTML, with Claude-Opus-4.6 at \$802.97 versus \$315.37 and DeepSeek-V4-Flash at \$6.75 versus \$1.91. In HTML the input cost is negligible and output dominates; in React the scaffold itself is billed as input tokens, so input cost often exceeds output cost. Dividing total cost by Total score, DeepSeek-V4-Flash achieves the best cost-per-point ratio in React, while Claude-Opus-4.7 costs roughly two orders of magnitude more per point. The mid-tier cluster---Mimo-V2-Pro, Kimi-K2.6, Qwen3.6-Plus, and GLM-5.1---occupies a pragmatic sweet spot where cost-per-point stays well below frontier levels without sacrificing substantial capability.

\vspace{-1mm}
\textbf{Failure attribution.} Most failures are infrastructural. All 856 HTML installation failures ($\approx$6.8\%) are verifier read-timeouts. In React, 79\% trace to port collisions or verifier-path issues; only 21\% are genuine code errors, dominated by two mechanically fixable patterns: Gemini-family models account for 35 of 37 syntax-level failures via unescaped backslashes in JS literals, and 12 ``phantom import'' errors reference non-existent \texttt{lucide-react} icons. The true failure ceiling is therefore lower than the nominal spread suggests.

\vspace{-1mm}
\textbf{Worked example.} Appendix~\ref{app:case_study} walks through a complete evaluation trace for a Super Mario query executed by Claude-Opus-4.7 in HTML mode. The example illustrates how static perception awards a provisional functionality score of 8.0 based on source-code completeness, while agent-driven interaction surfaces an emergent physics-tuning defect, the jump arc is too short to clear the first gap, that static inspection alone cannot detect. Stage~3 deferred scoring then adjusts the functionality score down to 7.0, demonstrating exactly the kind of embodied interaction gap \framename is designed to close.

\vspace{-1mm}
\textbf{Decomposition by dimension.} Figure~\ref{fig:cross-heatmap}, inAppendix~\ref{app:generation}, decomposes scores by language, difficulty, and L2 category. Medium tasks score highest (a non-monotonicity), while hard tasks fail on multi-step state management and easy tasks suffer from under-constraint. Tools~(Static) is the universal strength; Graphics and Animation the universal weakness. Language effects are weaker than difficulty effects, suggesting scaffold structure dominates over prompt language; see Appendix~\ref{app:generation} for detailed per-model breakdowns.

\vspace{-2mm}
\subsection{Ablation Study and Consistency Analysis}
\label{sec:ablation}
\vspace{-2mm}

\begin{table}[htbp]
\centering
\footnotesize
\setlength{\tabcolsep}{1.5pt}
\caption{Human agreement rates (\%) for ablated evaluation variants on 132 \dataname queries. Blue arrows show the drop relative to the full pipeline; the green arrow marks the single increase.}
\label{tab:ablation}
\newcommand{\downb}[1]{\textcolor{blue}{\tiny$\downarrow$#1}}
\newcommand{\upb}[1]{\textcolor{green!60!black}{\tiny$\uparrow$#1}}
\begin{tabular}{lccccccccc}
\toprule
& \multicolumn{3}{c}{\textbf{Dynamic}} & \multicolumn{3}{c}{\textbf{Static}} & \multicolumn{3}{c}{\textbf{Overall}} \\
\cmidrule(lr){2-4} \cmidrule(lr){5-7} \cmidrule(lr){8-10}
& Func. & Aesth. & Total & Func. & Aesth. & Total & Func. & Aesth. & Total \\
\midrule
w/o Vision & 30.7 \downb{23.1} & 52.4 \downb{2.2} & 55.1 \downb{5.9} & 26.3 \downb{24.1} & 51.0 \downb{3.5} & 50.3 \downb{12.7} & 29.5 \downb{23.4} & 52.0 \downb{2.6} & 53.8 \downb{7.8} \\
w/o Video & 33.0 \downb{20.8} & 29.6 \downb{25.0} & 56.4 \downb{4.6} & 29.6 \downb{20.8} & \textbf{55.0} \upb{0.5} & 57.5 \downb{5.5} & 32.1 \downb{20.8} & 36.5 \downb{18.1} & 56.7 \downb{4.9} \\
w/o Static Score & 53.6 \downb{0.2} & 52.7 \downb{1.9} & 58.6 \downb{2.4} & 47.4 \downb{3.0} & 50.1 \downb{4.4} & 55.6 \downb{7.4} & 52.0 \downb{0.9} & 52.0 \downb{2.6} & 57.8 \downb{3.8} \\
w/o Deferred Scoring & 38.5 \downb{15.3} & 29.6 \downb{25.0} & 41.2 \downb{19.8} & 26.5 \downb{23.9} & 23.8 \downb{30.7} & 39.1 \downb{23.9} & 35.2 \downb{17.7} & 28.0 \downb{26.6} & 40.6 \downb{21.0} \\
w/ Full pipeline & \textbf{53.8} & \textbf{54.6} & \textbf{61.0} & \textbf{50.4} & 54.5 & \textbf{63.0} & \textbf{52.9} & \textbf{54.6} & \textbf{61.6} \\
\bottomrule
\end{tabular}
\end{table}

We validate \framename against human judgment on 132 \dataname queries (12 per L2 category, 4 models: Qwen3.6-Plus, Mimo-V2-Pro, GLM-5.1, Claude-Opus-4.6), yielding 528 annotated instances. A pilot study revealed an instructive asymmetry: human raters find it easier to judge \emph{which} output is better (relative ranking) than to assign absolute scores to functionality or aesthetics. Consequently, we adopt different annotation protocols per dimension---\emph{pointwise} 8-point rubric scores for Functionality and Aesthetics, and a \emph{listwise} ranking for Total---and measure agreement at the pairwise level: both human and machine judgments are expanded into pairwise preferences, and a match counts when both agree on direction (or both tie). We also construct four ablated variants to isolate each evaluation stage: \textbf{w/o Vision} (source code only), \textbf{w/o Video} (adds one screenshot), \textbf{w/o Static Score} (drives interaction but skips static impression), and \textbf{w/o Deferred Scoring} (scores reactively per step).

\begin{figure*}[htbp]
\centering
\includegraphics[width=0.99\textwidth]{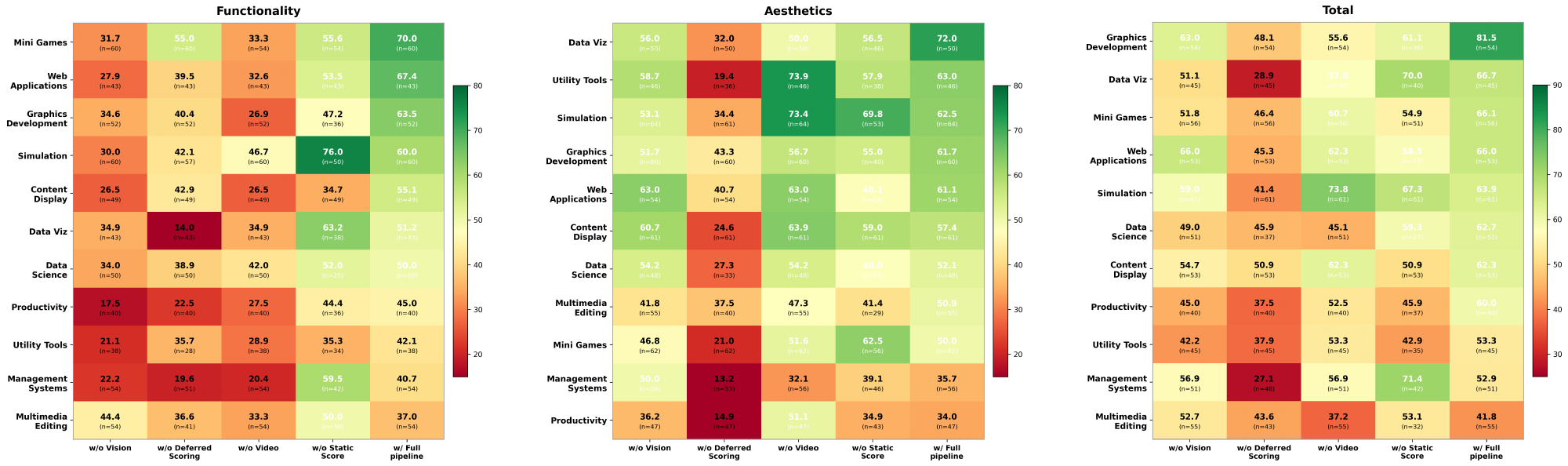}
\caption{Per-category human agreement rates (\%) for ablated evaluation variants on 132 queries. From left to right: Functionality, Aesthetics, and Total match rates. Darker green indicates stronger alignment with human judgment.}
\label{fig:ablation-heatmaps}
\vspace{-2mm}
\end{figure*}

Table~\ref{tab:ablation} reports aggregate human-agreement rates. The full pipeline achieves the highest average match rate at 61.6\%, outperforming w/o Vision by 7.8 points, w/o Video by 4.9 points, w/o Static Score by 3.8 points, and w/o Deferred Scoring by 21.0 points. It leads every split on Functionality (53.8 dynamic, 50.4 static), Aesthetics (54.6 dynamic, 54.5 static), and Total (61.0 dynamic, 63.0 static). The uniform dominance confirms that the three-stage design yields the most faithful proxy for human judgment.

\vspace{-1mm}
The absolute numbers may seem modest, but they sit well within the range reported for model-based evaluation of open-ended generation. Unlike closed-form tasks with gold answers, web development outputs vary enormously in implementation strategy, making inter-annotator agreement inherently lower. The ablation heatmaps in Figure~\ref{fig:ablation-heatmaps} further show that the gap between the full pipeline and ablated variants is consistent across almost all 11 categories, which strengthens the conclusion that the improvement is systematic rather than artifactual.

\vspace{-1mm}
A closer look at the category-level heatmaps reveals several instructive cross-dimensional inconsistencies. The largest Functionality--Aesthetics divergence occurs in \emph{Simulation} and \emph{Data Viz}: under w/o Video, Simulation scores 46.7\% on Functionality but 73.4\% on Aesthetics, because static screenshots often appear visually complete while hiding functional deficits (non-interactive physics, missing state transitions). Data Viz under w/o Deferred Scoring collapses to 14.0\% on Functionality yet stays at 32.0\% on Aesthetics, showing reactive step-by-step scoring fails to capture data-binding correctness. \emph{Management Systems} exhibits the largest Functionality--Total discrepancy: under w/o Static Score, Functionality reaches only 59.5\% while Total surges to 71.4\%, because listwise ranking tolerates partial functionality when the overall workflow feels coherent. Conversely, \emph{Productivity} under w/o Vision drops to 17.5\% on Functionality but remains at 45.0\% on Total, because relative ranking still recovers correct ordering even when pointwise scoring cannot verify tool behavior.

These patterns show that \framename's judgments mirror human meta-cognition: it struggles where humans struggle, namely distinguishing visual completeness from functional correctness, and succeeds where humans succeed, namely recognizing coherent workflows and relative quality, confirming that the verifier captures the same evaluative heuristics human annotators employ.

\vspace{-4mm}
\section{Conclusion}
\label{sec:conclusion}
\vspace{-3mm}
We presented \dataname, an 11-domain, 54-leaf, 1,000-query WebDev benchmark spanning three difficulty tiers and three target languages, together with \framename, a reference-free, autonomously driven evaluator that separates evidence accumulation from judgment across three stages. The benchmark results on 13 frontier LLMs reveal substantial headroom, particularly on dynamic interactive tasks and in the React agent-scaffold setting where component architecture imposes a higher floor for weaker models. The ablation study validates that each stage contributes measurably to alignment with human judgment, and the full pipeline achieves the highest agreement on every evaluated split. We hope \dataname and \framename serve the community as useful tools for automated evaluation and contribute to further developments in the field of web code generation.

\clearpage
\section*{Limitations}
\label{sec:limitation}
\dataname and \framename focus exclusively on front-end web generation and do not assess back-end components such as API design, database integration, authentication, or deployment, which are essential in production settings. The benchmark covers two generation modes, React agent scaffold and HTML direct chat, but generalization to other front-end frameworks or full-stack workflows remains future work. The human annotation set, while sufficient for validating the evaluator, is limited in scale (528 evaluated instances). Finally, language coverage is balanced across three target groups but does not exhaust the full multilingual web.

\section*{Boarder Impact}
\label{sec:boarder_impact}
\dataname and \framename reduce the cost of evaluating interactive web generation, enabling faster iteration on model architectures and broader stress-testing across 11 domains and multiple languages. Automated evaluators may, however, reward surface-level visual polish over accessibility and robustness, and cheaper evaluation could indirectly accelerate the development of systems capable of generating deceptive web content. We release the benchmark under terms that restrict commercial use for training generative models without safety review, and we urge the community to treat \framename as a complement to, rather than a replacement for, human oversight.

\section*{Declaration of LLM usage}
\label{sec:llm_usage}
The authors used large language models for writing assistance, including drafting, editing, and proofreading sections of this paper. All generated text was reviewed, revised, and fact-checked by the authors, who take full responsibility for the accuracy and integrity of the final content.


\begin{ack}
Use unnumbered first-level headings for the acknowledgments. All acknowledgments
go at the end of the paper before the list of references. Moreover, you are required to declare
funding (financial activities supporting the submitted work) and competing interests (related financial activities outside the submitted work).
More information about this disclosure can be found at: \url{https://neurips.cc/Conferences/2026/PaperInformation/FundingDisclosure}.

Do {\bf not} include this section in the anonymized submission, only in the final paper. You can use the \texttt{ack} environment provided in the style file to automatically hide this section in the anonymized submission.
\end{ack}



{
\small
\bibliographystyle{abbrvnat}
\bibliography{neurips_2026}

@misc{openai2026gpt54,
  author       = {{OpenAI}},
  title        = {Introducing {GPT-5.4}},
  year         = {2026},
  howpublished = {\url{https://openai.com/zh-Hans-CN/index/introducing-gpt-5-4/}},
  note         = {Accessed: 2026-04-23}
}

@misc{anthropic2026claudevisuals,
  author       = {{Anthropic}},
  title        = {{Claude} builds visuals},
  year         = {2026},
  howpublished = {\url{https://claude.com/blog/claude-builds-visuals}},
  note         = {Accessed: 2026-04-23}
}

@misc{google2026gemini3,
  author       = {{Google}},
  title        = {{Gemini 3}},
  year         = {2026},
  howpublished = {\url{https://aistudio.google.com/models/gemini-3}},
  note         = {Accessed: 2026-04-23}
}

@article{laurenccon2024unlocking,
  title={Unlocking the conversion of web screenshots into html code with the websight dataset},
  author={Lauren{\c{c}}on, Hugo and Tronchon, L{\'e}o and Sanh, Victor},
  journal={arXiv preprint arXiv:2403.09029},
  year={2024}
}

@article{sun2025fullfront,
  title={FullFront: Benchmarking MLLMs Across the Full Front-End Engineering Workflow},
  author={Sun, Haoyu and Wang, Huichen Will and Gu, Jiawei and Li, Linjie and Cheng, Yu},
  journal={arXiv preprint arXiv:2505.17399},
  year={2025}
}

@inproceedings{wu2024uiclip,
  title={UIClip: a data-driven model for assessing user interface design},
  author={Wu, Jason and Peng, Yi-Hao and Li, Xin Yue Amanda and Swearngin, Amanda and Bigham, Jeffrey P and Nichols, Jeffrey},
  booktitle={Proceedings of the 37th Annual ACM Symposium on User Interface Software and Technology},
  pages={1--16},
  year={2024}
}

@article{lai2025webrenderbench,
  title={WebRenderBench: Enhancing Web Interface Generation through Layout-Style Consistency and Reinforcement Learning},
  author={Lai, Peichao and Zhuang, Jinhui and Zhang, Kexuan and Xiong, Ningchang and Wang, Shengjie and Xu, Yanwei and Chen, Chong and Wang, Yilei and Cui, Bin},
  journal={arXiv preprint arXiv:2510.04097},
  year={2025}
}

@article{zhu2025frontendbench,
  title={Frontendbench: A benchmark for evaluating llms on front-end development via automatic evaluation},
  author={Zhu, Hongda and Zhang, Yiwen and Zhao, Bing and Ding, Jingzhe and Liu, Siyao and Liu, Tong and Wang, Dandan and Liu, Yanan and Li, Zhaojian},
  journal={arXiv preprint arXiv:2506.13832},
  year={2025}
}

@article{wan2024mrweb,
  title={Mrweb: An exploration of generating multi-page resource-aware web code from ui designs},
  author={Wan, Yuxuan and Dong, Yi and Xiao, Jingyu and Huo, Yintong and Wang, Wenxuan and Lyu, Michael R},
  journal={arXiv preprint arXiv:2412.15310},
  year={2024}
}

@article{chen2025iwr,
  title={IWR-Bench: Can LVLMs reconstruct interactive webpage from a user interaction video?},
  author={Chen, Yang and Liu, Minghao and Shen, Yufan and Li, Yunwen and Huang, Tianyuan and Fang, Xinyu and Zheng, Tianyu and Huang, Wenxuan and Yang, Cheng and Fu, Daocheng and others},
  journal={arXiv preprint arXiv:2509.24709},
  year={2025}
}

@article{xu2025webvia,
  title={Webvia: A web-based vision-language agentic framework for interactive and verifiable ui-to-code generation},
  author={Xu, Mingde and Yang, Zhen and Hong, Wenyi and Pan, Lihang and Fan, Xinyue and Wang, Yan and Gu, Xiaotao and Xu, Bin and Tang, Jie},
  journal={arXiv preprint arXiv:2511.06251},
  year={2025}
}

@article{lu2025webgen,
  title={Webgen-bench: Evaluating llms on generating interactive and functional websites from scratch},
  author={Lu, Zimu and Yang, Yunqiao and Ren, Houxing and Hou, Haotian and Xiao, Han and Wang, Ke and Shi, Weikang and Zhou, Aojun and Zhan, Mingjie and Li, Hongsheng},
  journal={arXiv preprint arXiv:2505.03733},
  year={2025}
}

@article{xu2025web,
  title={Web-bench: A llm code benchmark based on web standards and frameworks},
  author={Xu, Kai and Mao, YiWei and Guan, XinYi and Feng, ZiLong},
  journal={arXiv preprint arXiv:2505.07473},
  year={2025}
}

@misc{lei2026webcompassmultimodalwebcoding,
      title={WebCompass: Towards Multimodal Web Coding Evaluation for Code Language Models}, 
      author={Xinping Lei and Xinyu Che and Junqi Xiong and Chenchen Zhang and Yukai Huang and Chenyu Zhou and Haoyang Huang and Minghao Liu and Letian Zhu and Hongyi Ye and Jinhua Hao and Ken Deng and Zizheng Zhan and Han Li and Dailin Li and Yifan Yao and Ming Sun and Zhaoxiang Zhang and Jiaheng Liu},
      year={2026},
      eprint={2604.18224},
      archivePrefix={arXiv},
      primaryClass={cs.SE},
      url={https://arxiv.org/abs/2604.18224}, 
}

@inproceedings{yun2024web2code,
 author = {Yun, Sukmin and Lin, Haokun and Thushara, Rusiru and Bhat, Mohammad Qazim and Wang, Yongxin and Jiang, Zutao and Deng, Mingkai and Wang, Jinhong and Tao, Tianhua and Li, Junbo and Li, Haonan and Nakov, Preslav and Baldwin, Timothy and Liu, Zhengzhong and Xing, Eric P. and Liang, Xiaodan and Shen, Zhiqiang},
 booktitle = {Advances in Neural Information Processing Systems (NeurIPS)},
 pages = {112134--112157},
 publisher = {Curran Associates, Inc.},
 title = {Web2Code: A Large-scale Webpage-to-Code Dataset and Evaluation Framework for Multimodal LLMs},
 volume = {37},
 year = {2024}
}

@inproceedings{si2025design2code,
    title = "Design2Code: Benchmarking Multimodal Code Generation for Automated Front-End Engineering",
    author = "Si, Chenglei and Zhang, Yanzhe and Li, Ryan  and Yang, Zhengyuan  and Liu, Ruibo and Yang, Diyi",
    booktitle = "Proceedings of the 2025 Conference of the Nations of the Americas Chapter of the Association for Computational Linguistics: Human Language Technologies (NAACL)",
    month = apr,
    year = "2025",
    address = "Albuquerque, New Mexico",
    publisher = "Association for Computational Linguistics",
    pages = "3956--3974",
    ISBN = "979-8-89176-189-6"
}

@article{zhang2026miniappbench,
  title={MiniAppBench: Evaluating the Shift from Text to Interactive HTML Responses in LLM-Powered Assistants},
  author={Zhang, Zuhao and Yu, Chengyue and Li, Yuante and Zhuang, Chenyi and Mo, Linjian and Li, Shuai},
  journal={arXiv preprint arXiv:2603.09652},
  year={2026}
}

@article{he2026vision2web,
  title={Vision2Web: A Hierarchical Benchmark for Visual Website Development with Agent Verification},
  author={He, Zehai and Hong, Wenyi and Yang, Zhen and Pan, Ziyang and Liu, Mingdao and Gu, Xiaotao and Tang, Jie},
  journal={arXiv preprint arXiv:2603.26648},
  year={2026}
}

@article{xiao2025code,
  title={Code Aesthetics with Agentic Reward Feedback},
  author={Xiao, Bang and Jiang, Lingjie and Huang, Shaohan and Lv, Tengchao and Huang, Yupan and Wu, Xun and Cui, Lei and Wei, Furu},
  journal={arXiv preprint arXiv:2510.23272},
  year={2025}
}

@article{zhang2025artifactsbench,
  title={Artifactsbench: Bridging the visual-interactive gap in llm code generation evaluation},
  author={Zhang, Chenchen and Li, Yuhang and Xu, Can and Liu, Jiaheng and Liu, Ao and Zhou, Changzhi and Deng, Ken and Wu, Dengpeng and Huang, Guanhua and Li, Kejiao and others},
  journal={arXiv preprint arXiv:2507.04952},
  year={2025}
}

@article{liu2026webcoderbench,
  title={WebCoderBench: Benchmarking Web Application Generation with Comprehensive and Interpretable Evaluation Metrics},
  author={Liu, Chenxu and Fu, Yingjie and Yang, Wei and Zhang, Ying and Xie, Tao},
  journal={arXiv preprint arXiv:2601.02430},
  year={2026}
}

@article{kong2026webtestbench,
  title={WebTestBench: Evaluating Computer-Use Agents towards End-to-End Automated Web Testing},
  author={Kong, Fanheng and Zhang, Jingyuan and Yue, Yang and Sun, Chenxi and Tian, Yang and Feng, Shi and Yang, Xiaocui and Wang, Daling and Tian, Yu and Du, Jun and others},
  journal={arXiv preprint arXiv:2603.25226},
  year={2026}
}

@article{lu2026fullstack,
  title={FullStack-Agent: Enhancing Agentic Full-Stack Web Coding via Development-Oriented Testing and Repository Back-Translation},
  author={Lu, Zimu and Ren, Houxing and Yang, Yunqiao and Wang, Ke and Zong, Zhuofan and Zhan, Mingjie and Li, Hongsheng},
  journal={arXiv preprint arXiv:2602.03798},
  year={2026}
}

@article{qinghong2025computer,
  title={Computer-Use Agents as Judges for Generative User Interface},
  author={Lin, Kevin Qinghong and Hu, Siyuan and Li, Linjie and Yang, Zhengyuan and Wang, Lijuan and Torr, Philip and Shou, Mike Zheng},
  journal={arXiv preprint arXiv:2511.15567},
  year={2025}
}

@inproceedings{gui2025webcode2m,
  title={Webcode2m: A real-world dataset for code generation from webpage designs},
  author={Gui, Yi and Li, Zhen and Wan, Yao and Shi, Yemin and Zhang, Hongyu and Chen, Bohua and Su, Yi and Chen, Dongping and Wu, Siyuan and Zhou, Xing and others},
  booktitle={Proceedings of the ACM on Web Conference (WWW 2025)},
  pages={1834--1845},
  year={2025}
}

@article{jung2025ui,
  title={UI-Bench: A Benchmark for Evaluating Design Capabilities of AI Text-to-App Tools},
  author={Jung, Sam and Garcinuno, Agustin and Mateega, Spencer},
  journal={arXiv preprint arXiv:2508.20410},
  year={2025}
}

@inproceedings{xiao2025interaction2code,
  title={Interaction2code: Benchmarking mllm-based interactive webpage code generation from interactive prototyping},
  author={Xiao, Jingyu and Wan, Yuxuan and Huo, Yintong and Wang, Zixin and Xu, Xinyi and Wang, Wenxuan and Xu, Zhiyao and Wang, Yuhang and Lyu, Michael R},
  booktitle={2025 40th IEEE/ACM International Conference on Automated Software Engineering (ASE)},
  pages={241--253},
  year={2025},
  organization={IEEE}
}

@article{xiao2025designbench,
  title={Designbench: A comprehensive benchmark for mllm-based front-end code generation},
  author={Xiao, Jingyu and Wang, Ming and Lam, Man Ho and Wan, Yuxuan and Liu, Junliang and Huo, Yintong and Lyu, Michael R},
  journal={arXiv preprint arXiv:2506.06251},
  year={2025}
}

@article{gui2024vision2ui,
  title={Vision2ui: A real-world dataset with layout for code generation from ui designs},
  author={Gui, Yi and Li, Zhen and Wan, Yao and Shi, Yemin and Zhang, Hongyu and Su, Yi and Dong, Shaoling and Zhou, Xing and Jiang, Wenbin},
  journal={CoRR},
  year={2024}
}

@inproceedings{guo2025iw,
  title={Iw-bench: Evaluating large multimodal models for converting image-to-web},
  author={Guo, Hongcheng and Zhang, Wei and Chen, Junhao and Gu, Yaonan and Yang, Jian and Du, Junjia and Cao, Shaosheng and Hui, Binyuan and Liu, Tianyu and Ma, Jianxin and others},
  booktitle={Findings of the Association for Computational Linguistics: ACL 2025},
  pages={6449--6466},
  year={2025}
}

@misc{peng2026playcodermakingllmgeneratedgui,
      title={PlayCoder: Making LLM-Generated GUI Code Playable}, 
      author={Zhiyuan Peng and Wei Tao and Xin Yin and Chenhao Ying and Yuan Luo and Yiwen Guo},
      year={2026},
      eprint={2604.19742},
      archivePrefix={arXiv},
      primaryClass={cs.SE},
      url={https://arxiv.org/abs/2604.19742}, 
}

@techreport{sadowski2007simhash,
  title={Simhash: Hash-based similarity detection},
  author={Sadowski, Caitlin and Levin, Greg},
  year={2007},
  institution={Technical report, Google}
}

@article{robertson2004understanding,
  title={Understanding inverse document frequency: on theoretical arguments for IDF},
  author={Robertson, Stephen},
  journal={Journal of documentation},
  volume={60},
  number={5},
  pages={503--520},
  year={2004},
  publisher={Emerald Group Publishing Limited}
}

@article{flavell1979metacognition,
  title={Metacognition and cognitive monitoring: A new area of cognitive--developmental inquiry.},
  author={Flavell, John H},
  journal={American psychologist},
  volume={34},
  number={10},
  pages={906},
  year={1979},
  publisher={American Psychological Association}
}

@misc{claude-opus-4-7,
  author = {{Anthropic}},
  title = {Claude Opus 4.7},
  year = {2026},
  howpublished = {\url{https://www.anthropic.com/news/claude-opus-4-7}},
  note = {Accessed: 2026-05-01}
}

@misc{claude-opus-4-6,
  author = {{Anthropic}},
  title = {Claude Opus 4.6},
  year = {2026},
  howpublished = {\url{https://www.anthropic.com/news/claude-opus-4-6}},
  note = {Accessed: 2026-05-01}
}

@misc{gpt-5-4,
  author = {{OpenAI}},
  title = {Introducing {GPT}-5.4},
  year = {2026},
  howpublished = {\url{https://openai.com/index/introducing-gpt-5-4/}},
  note = {Accessed: 2026-05-01}
}

@misc{kimi-k2-6,
  author = {{Moonshot AI}},
  title = {Kimi {K}2.6},
  year = {2026},
  howpublished = {\url{https://www.kimi.com/blog/kimi-k2-6}},
  note = {Accessed: 2026-05-01}
}

@misc{glm-5-1,
  author = {{Zhipu AI}},
  title = {{GLM}-5.1},
  year = {2026},
  howpublished = {\url{https://z.ai/blog/glm-5.1}},
  note = {Accessed: 2026-05-01}
}

@misc{mimo-v2-pro,
  author = {{Xiaomi}},
  title = {Mimo {V}2 {P}ro},
  year = {2026},
  howpublished = {\url{https://mimo.xiaomi.com/mimo-v2-pro}},
  note = {Accessed: 2026-05-01}
}

@misc{deepseek-v4,
  author = {{DeepSeek AI}},
  title = {{DeepSeek}-{V}4},
  year = {2026},
  howpublished = {\url{https://huggingface.co/deepseek-ai/DeepSeek-V4-Pro/blob/main/DeepSeek_V4.pdf}},
  note = {Accessed: 2026-05-01}
}

@misc{gemini-3-1,
  author = {{Google DeepMind}},
  title = {Gemini 3.1 {P}ro},
  year = {2026},
  howpublished = {\url{https://deepmind.google/technologies/gemini/}},
  note = {Accessed: 2026-05-01}
}
}

\clearpage


\appendix

%

\section*{Appendix Contents}
\markboth{Appendix Contents}{Appendix Contents}
\addcontentsline{toc}{section}{Appendix Contents}

\begingroup
\small
\setlength{\parindent}{0pt}

\newcommand{\appsecentry}[2]{%
  \par\noindent
  \textbf{#1}\nobreak
  \leaders\hbox to 0.44em{\hss.\hss}\hfill
  \nobreak\pageref{#2}\par}
\newcommand{\appsubentry}[2]{%
  \par\noindent
  \hspace*{1.5em}#1\nobreak
  \leaders\hbox to 0.44em{\hss.\hss}\hfill
  \nobreak\pageref{#2}\par}

\vspace{0.4em}
\appsecentry{A.~\dataname Benchmark Data: Supplementary Details}{app:data}
\appsubentry{A.1 Query Deduplication Tool}{app:dedup_tool}
\appsubentry{A.2 LLM-as-Judge Quality Filtering}{app:llm_judge}
\appsubentry{A.3 Expert Review}{app:human_review}
\appsubentry{A.4 Task Taxonomy and Difficulty Rubric}{app:taxonomy}
\appsubentry{A.5 Difficulty Grading Rubric}{app:difficulty}
\appsubentry{A.6 Language Rebalancing Protocol}{app:lang_balance}
\appsubentry{A.7 Automated Classification Pipeline: Details and Error Analysis}{app:pipeline}

\vspace{0.6em}
\appsecentry{B.~\framename Evaluation Methodology: Supplementary Details}{app:method}
\appsubentry{B.1 Build, Deployment, and Interaction Details}{app:build_deploy}
\appsubentry{B.2 Interaction-Driving Prompt}{app:interaction_prompt}
\appsubentry{B.3 Judge-Agent Scoring Prompts}{app:judge_prompts}
\appsubentry{B.4 Human Annotation Rubric}{app:human_rubric}

\vspace{0.6em}
\appsecentry{C.~Worked Example: \dataname Evaluation Trace}{app:case_study}
\appsubentry{C.1 Query and Generated Output}{app:case_query}
\appsubentry{C.2 Stage 1: Static Perception}{app:case_static}
\appsubentry{C.3 Stage 2: Agent-Driven Interaction}{app:case_interact}
\appsubentry{C.4 Stage 3: Dynamic Scoring}{app:case_dynamic}
\appsubentry{C.5 What the Video Surfaced That Static Inspection Missed}{app:case_video}

\vspace{0.6em}
\appsecentry{D.~Detailed Generation Results}{app:generation}

\endgroup
\vspace{1.5em}

\section{\benchname Benchmark Data: Supplementary Details}
\label{app:data}

This appendix provides the full supplementary detail behind the \benchname construction described in Section~\ref{sec:data}, organized in the same order as the main text. Section~\ref{sec:data_sources} (Data Sources and Quality Assurance) is expanded by Appendix~\ref{app:dedup_tool} (deduplication tool), Appendix~\ref{app:llm_judge} (LLM-as-judge admissibility filter), and Appendix~\ref{app:human_review} (expert review). Section~\ref{sec:data_composition} (Data Composition and Statistics) is expanded by Appendix~\ref{app:taxonomy} (three-level task taxonomy), Appendix~\ref{app:difficulty} (six-dimension difficulty rubric with amplifier list), and Appendix~\ref{app:lang_balance} (stratified language rebalancing protocol). Section~\ref{sec:pipeline} (Automated Data Classification Pipeline) is expanded by Appendix~\ref{app:pipeline} (prompt template, validation protocol, error analysis, and revised leaf definitions). Each subsection is self-contained: definitions, algorithms, configurations, and auditable artefacts are reported in place rather than cross-referenced across subsections.

\subsection{Query Deduplication Tool}
\label{app:dedup_tool}

We developed an in-house query deduplication tool that operates entirely on local CPU, requires no online model dependency, and produces auditable merge traces. The tool is used for the first stage of the \benchname quality control pipeline (Section~\ref{sec:data_sources}) and is reusable for any downstream benchmark extension.

\paragraph{Motivation.}
Raw query pools sampled from production logs, social-media threads, and crowd authoring exhibit four recurrent pathologies: (i) verbatim duplicates inflating apparent dataset size; (ii) format variants that differ only in case, punctuation, whitespace, or trivial morphology; (iii) near-paraphrases with local synonymy, word reordering, or short rewrites that evade lexical matching; and (iv) opaque deduplication procedures that prevent downstream auditing of accidental merges. Purely manual cleaning is labor-intensive and subjective; purely neural approaches either depend on external embedding services or obscure the provenance of each merge.

\paragraph{Pipeline.}
The tool processes a CSV input through five deterministic stages.
\emph{(1) Normalization.} Each query is lower-cased, whitespace-collapsed, and stripped of edge punctuation to neutralize purely surface variation.
\emph{(2) Exact deduplication.} Post-normalization queries that collide are merged into a single representative, eliminating verbatim and format-variant repeats.
\emph{(3) Lexical near-duplicate pass (SimHash).} Each surviving query is hashed into a fixed-width SimHash fingerprint over character-level $n$-grams; pairs with Hamming distance below a configured threshold (default $3$ for $n{=}2$) are linked and clustered via connected components, absorbing minor edits, localized insertions or deletions, and short-range rewrites.
\emph{(4) Semantic near-duplicate pass (local TF-IDF).} A character $n$-gram TF-IDF representation is computed locally, and query pairs with cosine similarity above a tunable threshold (default $0.85$) are linked and merged into semantic clusters. This layer captures reordered phrasing, synonym substitutions, and short rewrites that are lexically distinct but intent-equivalent. The pass deliberately uses TF-IDF rather than a large neural encoder so that the entire pipeline remains reproducible offline, free of external API cost, and stable across environments.
\emph{(5) Representative selection.} Each cluster elects a single representative, and the tool emits both the deduplicated set and the full query-to-representative mapping.

\paragraph{Output artifacts.}
Every run produces four CSV files: \texttt{deduped\_queries.csv} (lexical representatives), \texttt{query\_groups.csv} (lexical cluster mapping), \texttt{semantic\_deduped\_queries.csv} (semantic representatives), and \texttt{semantic\_query\_groups.csv} (semantic cluster mapping). The dual-level output preserves both the clean output for downstream use and the provenance record for sample-level auditing, so that any suspected over-merging can be inspected and reversed without re-running the pipeline.

\paragraph{Operating statistics on \dataname.}
On the 1,000 raw candidates entering the quality pipeline, exact and SimHash passes reduce the set to 965, and the TF-IDF pass further removes one additional near-paraphrase cluster to yield 964 unique queries. The residual manual pass during expert review (Appendix~\ref{app:human_review}) is responsible for the remaining adjustments before the final 1,000-entry benchmark is admitted, with deliberately retained duplicates used as taxonomy-anchor controls.

\paragraph{Configuration and scope.}
For multilingual WebDev queries, we default to SimHash distance threshold $3$ at $n{=}2$ and TF-IDF cosine threshold $0.85$; both parameters are exposed to the user alongside column-name and output-path settings. The tool is intentionally scoped to a data-cleaning utility: it does not perform intent classification, long-document equivalence judgment, or deep semantic entailment, and is not a substitute for the LLM-judge and expert-review stages that follow.

\subsection{LLM-as-Judge Quality Filtering}
\label{app:llm_judge}

Queries that survive deduplication (Appendix~\ref{app:dedup_tool}) are screened by an LLM judge before entering expert review. This stage is designed to flag failure modes that are well-specified and objectively checkable, freeing expert reviewers to concentrate on judgment-heavy decisions (difficulty, scenario taxonomy, borderline scope).

\paragraph{Evaluation axes.}
Each candidate query is independently assessed along seven axes:
(i) \emph{safety}, screening for content that would be inadmissible regardless of task realism (harmful, illegal, or targeted content);
(ii) \emph{privacy leakage}, detecting personally identifiable information, credentials, or named non-public entities that require redaction;
(iii) \emph{task-direction consistency}, verifying that the query actually specifies a front-end web-development task rather than a backend, data-science, general-purpose coding, or non-coding request;
(iv) \emph{intent clarity}, identifying queries whose objective is ambiguous, internally contradictory, or under-specified to the point that no well-formed implementation is determinable;
(v) \emph{executability}, checking that the request can be realized as a self-contained front-end artifact runnable in a standard browser sandbox without proprietary services, live external APIs, or private datasets;
(vi) \emph{external-dependency minimality}, flagging queries whose fulfillment requires specific third-party URLs, paywalled assets, or user-specific credentials that would make reviewer reproduction infeasible; and
(vii) \emph{logical completeness}, ensuring that the query's functional specification does not contain mutually unsatisfiable constraints or dangling references.

\paragraph{Judgment protocol.}
The judge is prompted with the full taxonomy definitions, the seven criteria above, and a strict structured output format encoding per-axis verdicts and short rationales. Axes are evaluated independently to avoid correlated failures masking single-axis issues; a query is auto-rejected if any of (i), (ii), (v), or (vii) fires, auto-accepted if all seven pass, and routed to expert adjudication otherwise. We use a low-temperature configuration (matching the classification pipeline of Section~\ref{sec:pipeline}) to ensure reproducible verdicts across runs. To prevent single-model bias, a fraction of auto-accepted queries are re-judged by a second model during the expert-review stage as a consistency probe.

\paragraph{Observed failure distribution.}
Across the naturalistic pool, the dominant auto-rejection causes are external-dependency violations (queries referencing specific production URLs or private back-ends) and intent underspecification (single-phrase prompts with no operational definition). Across the crowd-synthesized pool, the dominant cause is task-direction drift (queries that, despite taxonomy anchoring, collapse into template CRUD patterns without principle-driven interaction) and, secondarily, logical incompleteness from over-aggressive expansion of seed intents. Routed (non-auto-decided) cases are resolved in expert review (Appendix~\ref{app:human_review}).

\paragraph{Scope.}
The LLM-judge stage is scoped to filtering for admissibility; it does not assign difficulty labels, does not determine taxonomy leaves (this is handled by the classification pipeline in Section~\ref{sec:pipeline}), and is not used as an arbiter of final benchmark quality. Its role is to reduce the expert-review burden to cases that genuinely require human judgment.

\subsection{Expert Review}
\label{app:human_review}

Queries surviving deduplication (Appendix~\ref{app:dedup_tool}) and LLM-judge filtering (Appendix~\ref{app:llm_judge}) enter a final expert-review stage before admission to \dataname. This stage is the sole authority for decisions that are inherently judgment-heavy and not reliably automatable: difficulty calibration, scenario-taxonomy placement at the leaf level, resolution of borderline admissibility cases routed by the LLM judge, and adversarial sampling of earlier automated decisions.

\paragraph{Reviewer pool.}
Review is conducted by annotators with front-end development background; each query is evaluated against the criteria below, and reviewers additionally mark queries that should be removed despite passing automated filters (e.g., queries whose phrasing is grammatical and safe but operationally trivial, or whose taxonomy placement cannot be uniquely determined even after definition refinement).

\paragraph{Review criteria.}
Expert review covers five decisions for each query:
(i) \emph{difficulty verification}, validating the difficulty level assigned from the six-dimension complexity schema (Appendix~\ref{app:difficulty}) and overriding when automated assignment conflicts with integrated reviewer judgment;
(ii) \emph{scenario-label verification}, spot-checking L1--L3 taxonomy labels produced by the classification pipeline (Section~\ref{sec:pipeline}), with particular attention to the three semantically adjacent pairs identified in the error analysis;
(iii) \emph{judge-decision audit}, re-examining a stratified sample of LLM-judge accept and reject decisions to catch systematic biases, with full overruling authority on individual cases;
(iv) \emph{executability confirmation}, verifying that at least one self-contained front-end implementation is feasible in a standard browser sandbox under the evaluation-time runtime constraints; and
(v) \emph{final admission}, the binary decision to admit the query to the benchmark or remove it.

\paragraph{Validation slice and inter-rater agreement.}
A fixed 540-query validation slice is annotated independently for the purposes of pipeline evaluation: the same slice is used to compute the 90.5\% human--model agreement of the classification pipeline (Section~\ref{sec:pipeline}), and to calibrate difficulty-label thresholds. Inter-rater agreement on the difficulty dimension over this slice is reported alongside the main results; disagreements are resolved by a third reviewer with taxonomy-definition authority.

\paragraph{Anonymization and revision.}
In addition to quality gating, expert review performs minimal query revision where required: removal of author-identifying markers and proprietary product names (replaced with generic descriptors), merging of multi-turn conversational threads into self-contained single-turn specifications, and light language normalization that preserves the original phrasing register (informal, colloquial, or code-switched) while removing artifacts of the source channel. No revision alters the functional requirements, success criteria, or difficulty of the query.

\paragraph{Outcome.}
The three-stage pipeline---automated deduplication, LLM-judge filtering, and expert review---admits 1,000 queries into the final benchmark. The pipeline is designed to be reusable: any future query pool added to \dataname passes through the same three stages, preserving comparability across dataset versions.

\subsection{Task Taxonomy and Difficulty Rubric}
\label{app:taxonomy}

Table~\ref{tab:taxonomy_static} and Table~\ref{tab:taxonomy_dynamic} present the complete three-level task taxonomy of WebDev.
Static web pages (Table~\ref{tab:taxonomy_static}) cover tasks whose content is determined at build time and requires no runtime state changes.
Dynamic web pages (Table~\ref{tab:taxonomy_dynamic}) cover tasks involving runtime user interaction, data mutation, or continuous state updates.

\begin{table*}[h]
\centering
\caption{Task taxonomy: \textbf{static web pages}. Each row represents a fine-grained task type (L3) under its functional domain (L2).}
\label{tab:taxonomy_static}
\resizebox{\textwidth}{!}{%
\begin{tabular}{llll}
\toprule
\textbf{Functional Domain (L2)} & \textbf{Task Type (L3)} & \textbf{Description} & \textbf{Example} \\
\midrule
\multirow{8}{*}{Display \& Content}
 & Corporate Website & Brand presentation, company info, contact details & Imitation of a brand homepage \\
 & Product Landing Page & Single product feature showcase, no transaction & Tesla model introduction page \\
 & Marketing Landing Page & Campaign-specific page with strong CTA & Double-11 promotion page \\
 & Blog / Article Page & Long-form reading with typography focus & Tech news article page \\
 & Documentation / Help Center & Structured technical docs with navigation & SaaS product manual \\
 & Event / Announcement Page & Event details, schedule, registration info & Tech conference page \\
 & Portfolio / Personal Page & Creative work showcase for individuals & Designer portfolio \\
 & Static Data Display & Pre-computed read-only data presentation & Annual revenue report page \\
\midrule
\multirow{6}{*}{Tools \& Productivity (Static)}
 & Unit Converter & Client-side unit conversion (length, weight, etc.) & Meter $\leftrightarrow$ Feet converter \\
 & Text Formatter & Code/text formatting and minification & JSON formatter / minifier \\
 & Encoder / Decoder & Encoding format conversion (Base64, URL, etc.) & Unicode converter \\
 & Validation Tool & Format/rule verification (regex, schema, etc.) & Regex online tester \\
 & Text Processing Tool & Batch text operations (dedup, case convert, etc.) & Text deduplication tool \\
 & Formula Calculator & Deterministic computation without external data & BMI calculator \\
\midrule
\multirow{6}{*}{Data Analysis \& Visualization (Static)}
 & Static Report Page & Fixed-period KPI/metric summary tables & Quarterly KPI report \\
 & Infographic / Data Story & Narrative data visualization on specific topics & Industry trend analysis page \\
 & Static Chart Page & Pre-rendered charts (bar, line, pie, map) & Regional sales bar chart \\
 & Fixed Dashboard / Display & Multi-metric display for presentations & Annual conference data screen \\
 & Static Comparison Page & Side-by-side data comparison & Plan effectiveness comparison \\
 & Data Archive Page & Historical report archive with navigation & Monthly report archive \\
\bottomrule
\end{tabular}%
}
\end{table*}

\begin{table*}[h]
\centering
\caption{Task taxonomy: \textbf{dynamic web pages}. Continued from Table~\ref{tab:taxonomy_static}.}
\label{tab:taxonomy_dynamic}
\resizebox{\textwidth}{!}{%
\begin{tabular}{llll}
\toprule
\textbf{Functional Domain (L2)} & \textbf{Task Type (L3)} & \textbf{Description} & \textbf{Example} \\
\midrule
\multirow{4}{*}{Interactive Applications}
 & Form \& Workflow & Multi-step forms with validation and submission & Registration / application flow \\
 & Content Interaction & Content consumption with like, comment, bookmark & Expandable-comment article page \\
 & Community \& Social & User-generated content, discussion, social features & Community discussion forum \\
 & Real-time Communication & WebSocket-based instant messaging and sync & Chat application \\
\midrule
\multirow{5}{*}{Management Systems}
 & Access \& Account Mgmt & User, role, permission, and org management & Enterprise permission backend \\
 & Business Object Mgmt & CRUD lifecycle for orders, products, customers & Order management system \\
 & Workflow \& Approval & Approval nodes, flow control, status tracking & Expense approval system \\
 & Configuration \& Rules & Visual business rule and parameter management & Business rule configuration \\
 & Data \& Resource Mgmt & Metadata governance, data lineage, asset catalog & Data asset management platform \\
\midrule
\multirow{3}{*}{Tools \& Productivity (Dynamic)}
 & Online File Processing & Server-side file conversion, merge, compression & PDF merge tool \\
 & Task \& Efficiency Tool & Time management, task tracking, habit building & Pomodoro timer \\
 & Real-time Info Tool & Live data query via external APIs & Real-time exchange rate tool \\
\midrule
\multirow{5}{*}{Multimedia \& Creative}
 & Image Editing & Crop, filter, text overlay, collage & Online photo editor \\
 & Video Editing & Cut, merge, transitions, subtitles & Online video editor \\
 & Audio Processing & Trim, denoise, speed change, format convert & Online audio editor \\
 & Rich Text Editing & Document editor with formatting and media & Online document editor \\
 & Visual \& Layout Design & Poster, presentation, social media design & Online PPT designer \\
\midrule
\multirow{6}{*}{Data Analysis \& Visualization (Dynamic)}
 & Interactive Dashboard & Filterable, drillable analytics dashboard & Operations analytics dashboard \\
 & Real-time Monitor & Live-updating monitoring with alerts & System status monitor screen \\
 & Self-service BI & Drag-and-drop data exploration and charting & Self-service BI platform \\
 & Interactive Report & Conditional query with pagination and export & Order query report \\
 & Experiment Analysis & A/B test results with statistical significance & A/B experiment analysis page \\
 & Data-driven Simulation & What-if analysis with parameter tuning & Business parameter simulator \\
\midrule
\multirow{4}{*}{Graphics Development}
 & 2D Graphics & Canvas/SVG drawing, algorithm visualization & Mermaid diagram / algo animation \\
 & 3D Graphics & WebGL/Three.js scene rendering and interaction & 3D model viewer \\
 & Visual Effects & Particle systems, shaders, dynamic animations & Particle system / shader effects \\
 & Interactive Graphics & Zoomable, pannable data/spatial exploration & Draggable/rotatable graphic view \\
\midrule
\multirow{4}{*}{Browser Games}
 & Puzzle \& Logic & Pattern recognition, strategy, level progression & Sudoku, jigsaw, match-3 \\
 & Action \& Reflex & Fast reaction, hand-eye coordination & Dodge, shoot, click-reaction \\
 & Strategy \& Simulation & Resource management, turn-based tactics & Turn-based / tower defense \\
 & Educational & Learning content embedded in game mechanics & Math / physics learning game \\
\midrule
\multirow{3}{*}{Simulation}
 & Physics Simulation & Parameter-driven scientific process simulation & Mechanics / EM simulation \\
 & Numerical Simulation & Mathematical model and system dynamics & Epidemic model / economic sim \\
 & Scenario Simulation & Multi-agent interaction and strategy evolution & RL environment / multi-agent \\
\bottomrule
\end{tabular}%
}
\end{table*}

\subsection{Difficulty Grading Rubric}
\label{app:difficulty}

\paragraph{Difficulty levels.}
Table~\ref{tab:difficulty_def} defines the four difficulty levels used in WebDev.
Each task is assessed along six orthogonal complexity dimensions (Table~\ref{tab:difficulty_rubric}), and the overall difficulty is determined by the highest dimension score.

\begin{table*}[h]
\centering
\caption{Difficulty level definitions.}
\label{tab:difficulty_def}
\resizebox{\textwidth}{!}{%
\begin{tabular}{lll}
\toprule
\textbf{Level} & \textbf{Definition} & \textbf{Typical Scope} \\
\midrule
Easy & Single-page, single-goal app with minimal logic (1--2 features) & Simple form, single-page site, basic display \\
Medium & App with a basic functional loop or clear interaction logic & Basic CRUD, multi-page website, simple game \\
Medium-Hard & Adds 1--2 complex modules (workflow, rules, multi-step state) & Filterable site, approval system, scoring game \\
Hard & Multi-module, multi-state, multi-role collaborative application & Complex platform, physics simulation, community site \\
\bottomrule
\end{tabular}%
}
\end{table*}

\begin{table*}[h]
\centering
\caption{Difficulty rubric: six complexity dimensions with level-specific criteria.}
\label{tab:difficulty_rubric}
\resizebox{\textwidth}{!}{%
\begin{tabular}{lllll}
\toprule
\textbf{Dimension} & \textbf{Easy} & \textbf{Medium} & \textbf{Medium-Hard} & \textbf{Hard} \\
\midrule
Functional Logic & 1--2 simple features & Basic functional loop & Multi-step, multi-state rules & Multi-module, rule-dense, complex boundaries \\
Page \& Interaction & Single page, short path & Multi-page or multi-region & Linked pages, long interaction chains & Multi-scene switching, complex state mgmt \\
Data \& System & No persistence or minimal & Basic CRUD / API calls & Multi-entity state, role separation & Multi-entity relations, permissions, notifications \\
Visual Design & Single-page display site & Multi-page site with structure & Forms, filtering, dynamic modules & Login, membership, payment, recommendations \\
User Experience & Basic display sufficient & Basic usability and visuals & High bar for feedback and guidance & Product-grade polish, sustained engagement \\
Dynamic Simulation & None or simple animation & Simple rule loops, basic motion & Multi-object linkage, basic collision & Gravity, friction, particles, complex state machines \\
\bottomrule
\end{tabular}%
}
\end{table*}

\paragraph{Difficulty amplifiers.}
Table~\ref{tab:amplifiers} lists 17 difficulty amplifiers---modular complexity factors that systematically escalate task difficulty when introduced.
These amplifiers serve as composable building blocks for constructing tasks at targeted difficulty levels.

\begin{table*}[h]
\centering
\caption{Difficulty amplifiers: modular complexity factors and their typical impact on difficulty level.}
\label{tab:amplifiers}
\resizebox{\textwidth}{!}{%
\begin{tabular}{clll}
\toprule
\textbf{\#} & \textbf{Amplifier} & \textbf{Description} & \textbf{Typical Escalation} \\
\midrule
1 & Approval Workflow & Node-based flow control with conditional branching & Medium $\rightarrow$ Medium-Hard / Hard \\
2 & Role-based Access Control & Identity-specific views, data scopes, and permissions & Medium $\rightarrow$ Medium-Hard / Hard \\
3 & State Machine & Object lifecycle with explicit state transitions & Medium $\rightarrow$ Medium-Hard \\
4 & Notifications & Async feedback, to-do lists, messaging mechanisms & Medium $\rightarrow$ Medium-Hard \\
5 & Charts \& Dashboards & Data aggregation, analysis, and visualization & Medium $\rightarrow$ Medium-Hard \\
6 & File Upload/Download & File handling, status tracking, and interaction flow & Medium $\rightarrow$ Medium-Hard \\
7 & External Service Integration & Maps, payment, AI, third-party API calls & Medium-Hard $\rightarrow$ Hard \\
8 & Sharing \& Viral Mechanics & Result sharing, invite links, social distribution & Medium $\rightarrow$ Medium-Hard \\
9 & Leaderboard / Points / Quests & Persistent state, competition, or growth mechanics & Medium $\rightarrow$ Medium-Hard / Hard \\
10 & Level Progression & Stage-based progression with evolving gameplay & Medium $\rightarrow$ Medium-Hard \\
11 & Collision / Gravity / Elasticity & Physics simulation capabilities & Medium-Hard $\rightarrow$ Hard \\
12 & Multi-object Real-time Interaction & Continuous mutual influence among multiple entities & Typically Hard \\
13 & Particle Effects / Continuous Rendering & Real-time visual effects with performance demands & Medium-Hard $\rightarrow$ Hard \\
14 & Login / Membership System & Account state, personal space, access control & Medium $\rightarrow$ Medium-Hard / Hard \\
15 & Content Management Backend & Upgrades display site to operable/manageable site & Medium $\rightarrow$ Medium-Hard \\
16 & Payment / Ordering & Transactional logic with order state management & Medium-Hard $\rightarrow$ Hard \\
17 & Community / Comments / Recommendations & Sustained interaction and content distribution & Medium-Hard $\rightarrow$ Hard \\
\bottomrule
\end{tabular}%
}
\end{table*}

\subsection{Language Rebalancing Protocol}
\label{app:lang_balance}

Naturally collected WebDev queries are overwhelmingly dominated by Chinese and English, with all remaining languages contributing a negligible tail. A benchmark that inherits this skew cannot cleanly separate genuine cross-lingual generalization from sheer language exposure in pretraining, and it offers no signal on the robustness of WebDev models to underrepresented languages. We therefore rebalance the language distribution of \dataname to a target split of 30\% Chinese, 30\% English, and 40\% distributed across six widely used additional languages, while explicitly preserving the joint distribution over $\text{L1}\times\text{L2}\times\text{difficulty}$ that the benchmark is designed to probe.

\paragraph{Target distribution.}
The raw pool of 1,000 queries after the quality pipeline contains 637 Chinese, 292 English, 49 Japanese, and a long tail of under-twenty-count languages. We set the rebalanced targets to 300 Chinese, 300 English, and around 66 queries each in French, Spanish, Japanese, German, Korean, and Portuguese, yielding 1,000 queries that exactly hit the 30 / 30 / 40 split. All languages outside these eight are redistributed.

\paragraph{Why cluster first.}
A naive strategy of randomly down-sampling Chinese and English to 239 each would almost certainly empty or heavily distort narrow strata such as high-difficulty static content-display pages. The design objective of \dataname is that capability gaps along one axis, such as language, should be measurable while holding the other axes constant, so rebalancing must leave the $\text{L1}\times\text{L2}\times\text{difficulty}$ shape intact. We therefore perform stratified subsampling within taxonomy-difficulty clusters, rather than flat random subsampling.

\paragraph{Stratification key.}
Each query is assigned a cluster key given by the tuple of its L1 label taking two values, L2 label taking eleven values, and difficulty taking three values, yielding up to 66 strata in total. Small strata containing only one or two queries are preserved without merging, since merging would itself introduce distribution drift.

\paragraph{Within-cluster stratified sampling.}
For each overrepresented language the global retention rate is fixed by the language's target over its current count, which gives approximately 55.2\% for Chinese and 81.8\% for English. Within each cluster containing $n$ queries of the source language, the retention quota is $n \times r$; we floor this value to obtain an integer lower bound, and then apply the largest-remainder method to assign the residual quota to the strata with the largest fractional remainders until the global target is matched exactly. Selection within each cluster is drawn uniformly at random under a fixed seed, so the rebalancing is fully reproducible.

\paragraph{Residual queries and target-language assignment.}
Queries not retained by the stratified sampling stage become the residual pool slated for translation into the six auxiliary languages. To satisfy the per-language quota of 53 entries, we construct a slot pool of labels sized to match the per-language deficit, for instance 49 French, 52 Spanish, 4 Japanese, 48 German, 51 Korean, and 52 Portuguese slots in our run, and we then randomly permute both the residual pool and the slot pool under the same seed and match them position-by-position. This global shuffling spreads target languages across strata rather than clustering any single language inside any single taxonomy or difficulty bucket, which is essential for holding the joint distribution constant.

\paragraph{Translation.}
Each residual query is re-translated from its original-language text into the assigned target language using a concurrent coding-oriented LLM endpoint. The translation prompt preserves code blocks, file paths, URLs, and structural markers such as role tags, section headings, and list bullets, and it translates only the natural-language portion of the query. Translations are cached to disk in ten-entry batches so that an interrupted run resumes without re-calling the API on already translated entries.

\paragraph{Distribution fidelity.}
After rebalancing, the language split matches the 30 / 30 / 40 target exactly, with 239 / 239 / 53 $\times$ 6. Because translation preserves L1, L2, and difficulty labels, and because stratified retention uses per-cluster proportional quotas, the marginal distributions over L1, L2, and difficulty are preserved up to stratum-level integer rounding. Every query retains an auditable trail recording its original language, final language, and the action taken, so that the rebalancing can be inspected or rolled back at the sample level.

\paragraph{Reproducibility.}
The rebalancing uses a single random seed for both the within-cluster retention draw and the global slot shuffle, and language detection uses a deterministic configuration. The translation cache, the rebalancing script, and the before-and-after summary sheet are released together with the benchmark.

\subsection{Automated Classification Pipeline: Details and Error Analysis}
\label{app:pipeline}

This subsection documents the full design and validation detail of the single-pass classification pipeline used to label all 1,000 \dataname queries.

\subsubsection{Pipeline Overview}
\label{app:pipeline_overview}

The pipeline operates as a stateless function $f:\text{query}\mapsto\text{L3 label}$.
Given a raw query $q$ (possibly multilingual and possibly containing attached code or UI references), the pipeline (i)~constructs a structured prompt $\mathcal{P}(q)$ that embeds the role specification, the full 85-category taxonomy with discriminative definitions, and an output-format contract;
(ii)~queries DeepSeek-V3.1 under near-deterministic decoding ($T{=}0.2$, $\mathrm{top}\text{-}p{=}0.9$);
and (iii)~parses the single-line JSON response \texttt{\{"task\_scenario": "<label>"\}} into an L3 label, from which the corresponding L2 and L1 are deterministically recovered via the taxonomy tree.
Malformed outputs are re-queried at most three times; in practice the retry rate is below $0.3\%$ and no query in the final benchmark required manual label arbitration due to decoder failure.

We deliberately restrict the pipeline to a single forward pass without self-consistency voting, chain-of-thought, or multi-model ensembling.
Two considerations motivate this choice.
First, near-deterministic decoding on a well-specified taxonomy already produces inter-run variance below $0.1\%$ (Table~\ref{tab:pipeline_val}), so additional sampling yields negligible accuracy gain.
Second, a single-pass formulation makes the pipeline cheap enough to re-run over the full benchmark whenever the taxonomy is revised, which is essential for the iterative definition-sharpening process described in Appendix~\ref{app:revised_defs}.

\subsubsection{Classification Prompt}
\label{app:prompt}

The prompt consists of four blocks.
\textbf{(i) Role specification.} The model is cast as a classification expert for Chinese--English (and mixed-language) web-development queries, with the explicit instruction to look past surface wording and infer the underlying technical scenario.
\textbf{(ii) Taxonomy definitions.} The full 85-category definition list is inlined verbatim; each entry pairs a category name with a one-paragraph definition that specifies the primary purpose, user-facing interactions, representative use cases, and explicit exclusions.
\textbf{(iii) Classification instructions.} The model is told to select \emph{exactly one} L3 label, with no multi-label assignment and no ``other'' option, and to base the decision on the \emph{task scenario} dimension rather than on the programming language or visual style that the query happens to mention.
\textbf{(iv) Output contract.} A strict JSON schema \texttt{\{"task\_scenario": "<label>"\}} with no surrounding commentary.
A trimmed representation of the prompt template is shown below.

\begin{tcolorbox}[
    enhanced,
    breakable,
    colback=white,
    colframe=black,
    title=\textbf{Prompt: Single-Pass Classification Pipeline (abridged)},
    fonttitle=\small,
]
\begin{Verbatim}[fontsize=\scriptsize, breaklines=true, breakanywhere=true]
You are an expert at classifying programming tasks in mixed
Chinese--English contexts. Look past surface wording and infer the underlying technical scenario.
Classify the query on a single dimension: task_scenario.
task_scenario (85 categories):
  - Corporate Homepage: <definition...>
  - Product Landing Page: <definition...>
  - ... (full 85-category list inlined verbatim) ...
  - Strategic Scenario Simulation: <definition...>

Query: {task}

Output format (strict):
{"task_scenario": "<one of the 85 labels>"}
No extra explanation. JSON only.
\end{Verbatim}
\end{tcolorbox}

\subsubsection{Validation Protocol and Results}
\label{app:validation}

We construct a held-out validation slice of 540 queries, stratified across L2 domains so that every functional domain contributes at least 12 instances.
Each query is independently labeled by two human annotators familiar with the taxonomy; disagreements are resolved by a third annotator, yielding a gold label set $G$.
We then run the pipeline three times with the same decoding configuration and distinct random seeds, producing prediction sets $\hat{Y}_1,\hat{Y}_2,\hat{Y}_3$, and report precision $\mathrm{Prec}_k=|\hat{Y}_k\cap G|/|G|$ on each run.

\begin{table}[h]
\centering
\small
\caption{Pipeline validation: three independent runs on the 540-query held-out slice.}
\label{tab:pipeline_val}
\begin{tabular}{lccc}
\toprule
\textbf{Run} & \textbf{TP} & \textbf{FP} & \textbf{Precision} \\
\midrule
Run 1 & 489 & 51 & 90.56\% \\
Run 2 & 488 & 52 & 90.37\% \\
Run 3 & 489 & 51 & 90.56\% \\
\midrule
Mean  & 488.7 & 51.3 & 90.50\% \\
\bottomrule
\end{tabular}
\end{table}

The near-zero run-to-run variance (standard deviation $<0.1\%$) confirms that low-temperature decoding yields highly stable labels on this taxonomy.
Errors are not uniformly distributed across the 85 leaves: fewer than $0.5\%$ of errors occur on category pairs that are semantically well-separated (e.g.\ \emph{Mini Game} vs.\ \emph{Corporate Homepage}), and roughly $80\%$ of all errors concentrate on the three category pairs analyzed below.

\subsubsection{Error Analysis and Revised Definitions}
\label{app:revised_defs}

We manually audited all 51--52 false positives across the three runs and clustered them by the (gold, predicted) pair.
Three pairs account for the majority of errors; for each we report the original definitions, representative bad cases, the root cause, and the revised definitions that were adopted for the final benchmark.

\paragraph{Static Data Report vs.\ Static Data Display.}

\textit{Original definitions.}
\emph{Static Data Report}: ``a read-only page whose purpose is to summarize business data and KPI indicators for a fixed time period, presented as tables and metric cards.''
\emph{Static Data Display}: ``a read-only page presenting pre-computed data, statistical results, or business metrics, with content fixed at publication time.''

\textit{Root cause.}
Both definitions emphasize ``read-only'' and ``fixed data''; when the query mentions KPI reports, inventory reports, or status reports, both definitions superficially match, and the model defaults to the broader \emph{Static Data Display} as a fallback.

\begin{table}[h]
\centering
\small
\caption{Representative bad cases: Static Data Report vs.\ Static Data Display.}
\label{tab:bad_case_report}
\begin{tabular}{p{9cm}ll}
\toprule
\textbf{Query} & \textbf{Gold} & \textbf{Predicted} \\
\midrule
Build an annual KPI attainment report page showing department-level metric completion and trends & Report & Display \\
Build a static inventory report page with stock levels by category & Report & Display \\
Build an ad-campaign performance report page showing impressions, click-through rate, and conversion rate & Report & Display \\
Design a project status report page with milestone completion and resource usage & Report & Display \\
\bottomrule
\end{tabular}
\end{table}

\textit{Revised definitions.}
\emph{Static Data Report} is tightened to require an \emph{explicit recurrence period} (daily/weekly/monthly/quarterly/annual), a defined KPI dimension set, and a structured layout with report title, period range, summary metric cards, dimension breakdown tables, and trend comparison charts.
Canonical examples: monthly sales report, quarterly KPI dashboard, annual financial report, project weekly status report.
\emph{Static Data Display} is restricted to \emph{non-periodic, one-off} data releases such as single-year report data pages, fixed public-indicator boards, or research-findings release pages.

\paragraph{Static Comparison Page vs.\ Experiment Analysis.}

\textit{Original definition.}
\emph{Static Comparison Page}: ``a page that compares data across two or more objects/plans/time periods using tables or comparison charts, with data fixed at publication time.''

\textit{Root cause.}
The concept of ``comparison'' is implicitly shared across multiple labels, including \emph{Experiment Analysis} for A/B tests, \emph{Product Landing Page} for specification comparison, and \emph{Marketing Page} for channel effectiveness comparison.
When the query mentions ``A/B plan comparison'' or ``product comparison,'' the model is triggered by the surface cue rather than by the underlying distinction between \emph{static result presentation} and \emph{live experimentation}.

\begin{table}[h]
\centering
\small
\caption{Representative bad cases: Static Comparison vs.\ Experiment Analysis and adjacent labels.}
\label{tab:bad_case_compare}
\begin{tabular}{p{9cm}ll}
\toprule
\textbf{Query} & \textbf{Gold} & \textbf{Predicted} \\
\midrule
Build an A/B plan comparison page showing conversion rate and revenue for the two plans & Comparison & Experiment \\
Build a product specification comparison page for different model variants & Comparison & Product Page \\
Create a marketing channel ROI comparison showing performance by channel & Comparison & Marketing Page \\
\bottomrule
\end{tabular}
\end{table}

\textit{Revised definition.}
\emph{Static Comparison Page} is a read-only page presenting the results of a \emph{completed} multi-object, multi-plan, or multi-period comparison; the data is fixed at publication, and the page does not support initiating new analyses or experiments.
Canonical examples: competitive-product analysis reports, plan-evaluation comparison tables, historical year-over-year or month-over-month displays, technology-selection comparison documents.

\paragraph{Data-driven Simulation vs.\ Numerical Simulation.}

\textit{Original definitions.}
\emph{Data-driven Simulation}: ``a page whose core function is business-scenario modeling, parameter tuning, and what-if analysis.''
\emph{Numerical Simulation}: ``a page based on mathematical models, dynamical systems, or numerical methods.''

\textit{Root cause.}
Both definitions permit parameter adjustment and real-time response visualization.
Queries such as ``pricing strategy simulator,'' ``demand forecasting tool,'' or ``Monte Carlo dashboard'' simultaneously involve mathematical computation \emph{and} business decision support, so the model cannot choose between them on surface cues alone.

\begin{table}[h]
\centering
\small
\caption{Representative bad cases: Data-driven Simulation vs.\ Numerical Simulation.}
\label{tab:bad_case_sim}
\begin{tabular}{p{9cm}ll}
\toprule
\textbf{Query} & \textbf{Gold} & \textbf{Predicted} \\
\midrule
Design a demand forecasting tool with parameter sensitivity analysis & Data-driven & Numerical \\
Build a pricing-strategy simulator that adjusts price-elasticity parameters to predict sales and revenue & Data-driven & Numerical \\
Build a capacity planning simulator with resource allocation scenarios & Data-driven & Numerical \\
Design a Monte Carlo simulation dashboard for risk assessment & Data-driven & Numerical \\
\bottomrule
\end{tabular}
\end{table}

\textit{Revised definitions.}
\emph{Data-driven Simulation} is redefined as a \emph{business decision-support} simulation tool targeted at non-technical users (business analysts, product managers, operations staff), who adjust business parameters (price, budget, inventory level) to predict business outcomes (revenue, ROI, sales volume).
The emphasis is on the ``business assumption $\rightarrow$ business outcome'' causal chain rather than on the validity of the underlying mathematical model.
\emph{Numerical Simulation} is redefined as a \emph{scientific or engineering} simulation page targeted at researchers, engineers, and students, used to understand and analyze the mathematical properties of complex systems such as differential equation solving, Monte Carlo sampling, and dynamical system evolution.

\subsubsection{Post-revision Effective Precision}

After adopting the revised definitions, we re-ran the pipeline on the same 540-query validation slice.
The three systematic error pairs above no longer dominate the residual error set, and the effective precision on the final benchmark labels is estimated to exceed $95\%$.
Because the benchmark-release labels were additionally spot-checked by human annotators, any remaining disagreements were resolved in favor of the human judgment and the corresponding definitions were logged for future revision.

\section{\dataname Evaluation Methodology: Supplementary Details}
\label{app:method}

This appendix provides the full supplementary detail behind the \framename evaluation methodology described in Section~\ref{sec:eval_method}, organized in the same order as the main text. Section~\ref{sec:eval_framework} (Agent-Driven Interactive Evaluation) is expanded by Appendix~\ref{app:build_deploy} (standardized build-and-deploy pipeline), Appendix~\ref{app:interaction_details} (three agent interaction mechanisms: environment freeze, multi-modal evidence capture, and human-like input simulation), Appendix~\ref{app:interaction_prompt} (the Stage-2 driving prompt used by \framename's computer-using agent), and Appendix~\ref{app:judge_prompts} (the full judge-agent prompts that fix the reviewer-side scoring priors shared across all queries). Section~\ref{sec:eval_dims} (Evaluation Dimensions) is expanded by Appendix~\ref{app:human_rubric} (sixteen-item human rubric with design principles, scene-adapted criteria, exemption rules, and the aggregation formula from binary items to the two reported dimensions). Each subsection is self-contained: definitions, protocols, and auditable artifacts are reported in place rather than cross-referenced across subsections.

\subsection{Build, Deployment, and Interaction Details}
\label{app:build_deploy}

\paragraph{Standardized build and deployment.}
Before evaluation, each generated codebase is passed through a standardized build-and-deploy pipeline to ensure a fair and reproducible environment.
The pipeline automatically resolves dependencies, compiles the project, and validates the build output; samples that fail to produce a valid artifact are recorded as build failures and excluded from subsequent evaluation stages.
Successfully built applications are deployed as live, browser-accessible instances that serve as the direct evaluation target, closely mirroring the conditions under which real end-users would access the application.

\paragraph{Generation scaffold.}
\label{app:scaffold}
The React setting provides an empty Vite-based React project with Tailwind CSS and shadcn/ui. The model modifies this scaffold through tool calls to implement the query, generating repo-level code that must be built and installed. The scaffold supports both static and dynamic pages equally; the model decides the appropriate component architecture for each query. We do not consider scaffold-based HTML generation in this work; all scaffold runs produce React projects. In the HTML setting, the model receives only the user query in a standard chat interface and is instructed to output a single self-contained HTML file without any scaffold or build pipeline.

The scaffold exposes a tool set covering file operations (create, read, edit, delete, list, glob, grep, patch), project execution (build, npm install), web access (search, fetch), image generation, and plan updates. File creation, reading, editing, and regex search are available in both modes; deletion, listing, glob matching, project building, dependency installation, and image generation are scaffold-only. Patch application and plan updates are reserved for Codex.

\label{app:interaction_details}
\paragraph{Agent interaction mechanisms.}
To support reliable evaluation under dynamic web environments, we introduce three modifications to the standard agent interaction loop:

\begin{enumerate}[nosep,leftmargin=*]
  \item \textbf{Environment freeze.} During the agent's deliberation phase (between observation and action selection), the application state is paused to prevent temporal drift. This ensures that the page state the agent reasons about remains consistent with the state it subsequently acts upon, avoiding evaluation artifacts caused by animations, timers, or asynchronous updates that advance while the agent deliberates.

  \item \textbf{Multi-modal evidence capture.} A continuous capture pipeline records screen video and audio streams alongside per-step screenshots throughout the interaction session. Unlike discrete snapshot-based approaches, this preserves the full temporal evolution of application behavior, including animation timing, transition smoothness, loading-state flicker, and audio feedback, providing the dynamic scoring stage with evidence that would otherwise be lost between observation points.

  \item \textbf{Human-like input simulation.} Rather than issuing instantaneous programmatic inputs, the agent introduces realistic interaction rhythms: gradual mouse movements, natural typing cadence, and appropriate pauses between actions. This prevents evaluation artifacts that arise when applications behave differently under programmatic versus human-speed input (e.g., debounce-guarded controls, hover-triggered tooltips, drag interactions with velocity-dependent behavior).
\end{enumerate}

%

\subsection{Interaction-Driving Prompt}
\label{app:interaction_prompt}

This appendix reproduces, verbatim, the system prompt that drives \framename's Stage~2 (\emph{Agent-Driven Interaction}, Section~\ref{sec:eval_framework}): the computer-using agent that autonomously explores the deployed application and records the multi-modal evidence package that Stage~3 later consumes. The prompt in this appendix is orthogonal to the scoring prompts of Section~\ref{app:judge_prompts}: it governs \emph{how the agent drives the page}, whereas those govern \emph{how the judge assigns scores}.

Two properties of this prompt matter for \framename's reference-free claim. First, it is \emph{fixed once} and \emph{reused across every query} on \dataname; the only per-query input is supplied through placeholders (\texttt{\{url\}}, \texttt{\{task\_prompt\}}, \texttt{\{max\_steps\}}) that carry the deployment handle, the user's original brief, and a global step budget---no reference implementation, expected interaction trajectory, or target completion state is ever injected. Second, the prompt explicitly instructs the agent to \emph{interact and document, not to evaluate or score}; the agent's output is a behavioral trace and a neutral observation summary, and any quality verdict is deferred to the scoring stages. These two properties ensure that the driving stage remains an evidence-collection step rather than a hidden judgment step.

\begin{tcolorbox}[
    enhanced,
    breakable,
    colback=white,
    colframe=black,
    title=\textbf{Prompt: Stage-2 Interaction Driving (\framename's Computer-Using Agent)},
    fonttitle=\small,
]
\begin{Verbatim}[fontsize=\scriptsize, breaklines=true, breakanywhere=true]
Go to the website at {url}.

IMPORTANT -- DIRECT NAVIGATION REQUIREMENT:
- You MUST directly navigate to this exact URL.
- Do NOT use a browser search engine.
- Do NOT open search results, previews, cached pages, or redirects.
- You must load and interact with ONLY the webpage served at the given URL.

You are an expert web application tester. Your task is to interact with the application,
test its features, and document your interaction journey.

You are NOT required to evaluate or score the application -- just interact and report what
happened.

I. CORE OPERATING PRINCIPLES

1. Primacy of Evidence: Document every action and its directly observed result. If no
   visible change occurred, say so explicitly.
2. Anti-Stagnation: If the page state is identical for 3 consecutive observation cycles
   while waiting for something (e.g., a response, animation, or load), note this as "no
   change observed" and move on.
3. Anti-Loop: If you find yourself repeating the same action or sequence without a new
   outcome, break the loop immediately and document it.
4. Garbled Page Handling: If the page appears garbled or unreadable (encoding issues),
   refresh up to 2 times. If still broken, exit immediately and report it in your
   overall_observation -- do not attempt workarounds.
5. Programming Language: The output project is expected to be a web application (HTML,
   React, etc.), regardless of what the query specifies. Do NOT document or comment on any
   programming language mismatch -- just test what is in front of you.
6. Page Language: Note in your overall_observation if the page language does not match the
   language requirement in the user query.
7. No Scoring Required: You are NOT evaluating quality -- just testing functionality and
   documenting observations.

II. WEB APPLICATION TESTING PROTOCOL

Step 1: Test Core Functionality
Test the CORE LOGIC first. Interact with the main workflow users will use the website, and
experience its primary purpose.

Step 2: Test Interactive Elements
Systematically test:
1. Navigation: All menu items, links, breadcrumbs
2. Buttons: All primary and secondary buttons
3. Forms: Input fields, dropdowns, checkboxes, validation
4. Dynamic Content: Tabs, accordions, modals, tooltips
5. Media: Images, videos, carousels load status
6. Search/Filter: Any search or filtering functionality

III. GAME TESTING PROTOCOL

If the application appears to be a game, apply the following protocol instead of (or in
addition to) the Web Application Testing Protocol above.

Non-Real-Time Interactive Games (card games, turn-based games, strategy games, puzzle games,
etc.):
- Try to test the complete game loop: start -> interact -> win/lose -> restart/next
- Test Additional Features:
  - Settings/options (if available)
  - Instructions/help (if available)
  - Any other interactive elements (e.g. buttons)

Real-Time Interactive Games (Snake, Flappy Bird, racing games, shooting games,
Minecraft-like games, etc.):
- Try to verify that the game's core controls work
- Do NOT try to complete the game, test all features, or achieve a high score -- just test
  core interactions

Special Notes for All Games:
- MOST games require holding a key to move/fly, not just pressing it once; you can press
  the key more times instead.
- Some games require clicking the canvas to lock the cursor before controls work
  (especially Minecraft-like games).
- Some operations are only possible under specific conditions (e.g. breaking a block
  requires being close enough in Minecraft-like games) -- avoid these if they are hard to
  reach.

IV. SAMPLING STRATEGY

For applications with many similar elements:
- Repeated Elements: Test 3-5 representative samples (e.g., if 20 product cards, test 3-5).
- Document Sampling: Note in summary like "Tested 3/20 product cards, all functional".
- Focus on Variety: Test different types of elements rather than all instances of the same
  type.

V. TASK DESCRIPTION

{task_prompt}

VI. EXECUTION REQUIREMENTS

1. Step Limit: You MUST conclude and produce your final JSON output within {max_steps}
   steps. Actively plan your steps, prioritize core functionality, and stop testing in time
   to prepare output before reaching the limit.
2. Environment Limitations (DO NOT Test -- Assume Working):
   - Backend/Database: Login systems, user authentication, data persistence
   - Third-Party APIs: LLM APIs, payment gateways, social media APIs, map services
   - File Operations: File upload/download functionality
   - Email/SMS: Email sending, SMS verification
3. Console Errors -- Only Report Critical Errors: Record only JavaScript errors that cause
   observable malfunction in the page or directly break functionality. Do NOT report:
   - Console warnings of any kind
   - Font loading failures (fonts.googleapis.com, fonts.gstatic.com)
   - Favicon 404 errors
   - CDN resource failures that do not visibly break the page
4. Focus on Frontend Interactions: Test what's visibly interactive in the browser.
5. Prioritize Critical Path: Test main workflow first, then secondary elements.
6. No Edge Cases: Do NOT test extreme or edge case inputs.

VII. FINAL OUTPUT FORMAT

Provide a comprehensive interaction summary as a JSON object. This summary will be used by
a VLM evaluator to assess the application quality.

Important: You MUST provide this JSON output before reaching the {max_steps}-step limit.

{
  "actions_performed": [
    "Navigated to {url}",
    "Clicked navigation menu item '...'",
    ...
  ],
  "console_errors": [
    "Critical error: TypeError: Cannot read property 'x' of undefined at ... -- caused
     submit button to fail",
    "(Leave empty list [] if no critical errors were observed, do NOT include non-critical
     errors)"
  ],
  "overall_observation": "Comprehensive summary of your testing experience. Describe what
   worked correctly, what had problems, any garbled/encoding issues, page language
   mismatches, or features that could not be tested due to environment limitations. Note
   any limitations due to automation constraints or step limits."
}

Remember: Your role is to INTERACT and DOCUMENT, not to evaluate or score. The VLM will
handle evaluation based on your interaction trace and the screenshots.
\end{Verbatim}
\end{tcolorbox}

%
%

\subsection{Judge-Agent Scoring Prompts}
\label{app:judge_prompts}

This appendix reproduces, verbatim, the three prompts that drive the judge agent across \framename's scoring stages. Two properties of these prompts are worth naming explicitly. First, all three are \emph{fixed} before any query is drawn and \emph{shared across every query} on \dataname; they encode reviewer-side scoring priors—what ``complete,'' ``broken,'' or ``polished'' mean in the abstract—rather than per-query oracles derived from a reference implementation. Second, no prompt injects a task-specific checklist, expected output, or correctness trace; the judge still has to reason over the live evidence package it is handed. In that sense the prompts play the same role as a seasoned reviewer's internalized standards: calibration, not an answer key.

Prompt~\ref{app:judge_prompts:static} is invoked at Stage~1 (\emph{Static Perception}) on a single rendered frame plus source code, console logs, and the original query, yielding a provisional pair of functionality and aesthetics scores. Prompts~\ref{app:judge_prompts:video_detect} and~\ref{app:judge_prompts:video_adjust} are invoked at Stage~3 (\emph{Dynamic Scoring}): the first mines the agent's interaction video and screen recording for defects not visible in a static frame; the second combines those video-surfaced defects with the Stage~1 scores to produce the final calibrated scores with structured failure attribution.

\subsubsection{Stage 1 --- Static Scoring Prompt}
\label{app:judge_prompts:static}

\begin{tcolorbox}[
    enhanced,
    breakable,
    colback=white,
    colframe=black,
    title=\textbf{Prompt: Static Perception Scoring (Stage 1)},
    fonttitle=\small,
]
\begin{Verbatim}[fontsize=\scriptsize, breaklines=true, breakanywhere=true]
You are an expert UI/UX designer and code reviewer. Evaluate the provided webpage on both
aesthetics and functionality based on the provided screenshot, source code, user query, and
runtime logs.

I. INPUT DATA
1. Webpage Screenshot: For visual and layout audit.
2. Source Code: To review logic, event handlers, and implementation quality.
3. Original User Query: To verify if requirements and language match.
4. Browser Console Logs & Dev Server Output: To detect hidden functional crashes or warnings.

II. SCORING SYSTEM (0.0 - 8.0 SCALE)

Programming language specification: The output project MUST be in html/react WHATEVER the
query specified. DO NOT DEDUCT points for the difference from user query.
Page language check: The web page's displayed language (in contrary to the programming
language) MUST follow the query requirements; if not specified, it should be the same as the
user query. A language mismatch is a significant quality problem and should be reflected in
the functional score.

1. FUNCTIONAL SCORING (REQUIREMENT-DRIVEN AUDIT)

Step 1: Assess Implementation Completeness
- 8.0 points: ALL user requirements fully implemented with correct logic
- 7.0 points: ALL user requirements implemented, only minor features missing or error (e.g.
  form validation is incomplete)
- 5.0 points: Core requirements implemented, some key features missing or error (e.g. game
  score calculation error; a simulator without updates when user settings change)
- 3.0 points: Core logic incomplete (e.g. a game that does not have correct
  "start->interact->win/lose->restart/next" loop; a booking system that cannot reach the
  final step)
- 1.0 points: Code provides only a skeleton; all interactions are missing or logically
  erroneous (e.g. a static page without any required interactions)
- 0.0 points: Page fails to render, is blank, shows raw code or placeholder page (e.g.
  "ready to build"), or completely irrelevant to query

Step 2: Verify Instruction Following & Query-Output Alignment
Check the output against the original user query for these problems:
- Query-output alignment: If the query asks for a tool, game, calculator, editor, or any
  functional application, the output must BE that functional application -- not a marketing
  page, landing page, or product showcase that merely describes or wraps the tool. A landing
  page for a tool is NOT the tool itself.
- Reasonable elaboration: The model may add features beyond the query, but they must be
  contextually reasonable. Adding payment plans, subscription tiers, or pricing sections to
  a personal portfolio, school project, simple utility, or any context where monetization
  makes no sense is unreasonable elaboration and should lower the score.
- No hallucinated content: The output should not contain invented brand names, product
  names, company names, or organization names that are not present in the query unless they
  are contextually necessary.
- Template detection: If the page is essentially unchanged from default starter boilerplate
  (e.g., default Create React App page, generic SaaS template with stock sections), it
  should score very low (1.0-2.0) regardless of whether it technically "builds" and
  "renders."
- Language mismatch: If the displayed page language does not match the query language (or
  explicit language requirements), this is a significant problem that should lower the score
  meaningfully.

Step 3: Verify Interactive Elements via Source Code
For EACH interactive element (button/link/form) mentioned in the user requirements:
- Check the source code directly -- do not rely solely on visual inspection of the
  screenshot.
- Look for buttons/links in JSX that have no onClick handler, empty handler functions, or
  handlers that only call alert() or console.log().
- Check if form submissions actually process data (update state, call APIs, produce results)
  rather than just preventing default.
- Verify that data-display components are wired to real state/data, not hardcoded
  placeholder values.
- Placeholder or non-functional interactive elements should lower the score proportionally
  to their importance (critical actions like submit/save/play are more damaging than minor
  UI toggles).

Step 4: Check Data Display Quality
- All data clearly visible and readable: No issue
- Minor readability issues (small font, low contrast): Minor problem
- Element overlapping or data completely unreadable: Significant problem per instance

Step 5: Check Console Errors
- Critical console ERRORs that cause observable malfunction in the screenshot, affect
  functionality, or indicate broken code logic should lower the score. Only deduct once if
  the same problem causes continuous error throwing.
- Non-Critical Errors (NO impact): External resource loading failures that have graceful
  fallback and do not break functionality, including Fonts API failures
  (fonts.googleapis.com, fonts.gstatic.com) and 404 for favicon.ico loading.
- Console WARNINGS are acceptable and NOT penalized.

Final Functional Score: Synthesize findings from all steps above into a single score on the
0.0-8.0 scale. Explain your reasoning step by step with a clear calculation breakdown.

2. AESTHETIC SCORING (DEFECT-BASED ELITE STANDARDS)

STEP 1: BASELINE
- 5.0 points: High-quality, clean, and modern. Standard professional work.
- 3.0 points: Functional but unpolished. With issues.
- 1.0 points: Raw HTML elements with no styling or only very basic CSS.
- 0.0 points: App fails to render, is blank, shows raw code or placeholder page (e.g. "ready
  to build"), visually broken, or is irrelevant to the query.

STEP 2: Evaluate for Bonus Quality (Evidence Required to show it's EXCELLENT, not just HAS)
- Outstanding visual hierarchy (font size/weight/color/spacing variation that clearly guides
  the eye)
- Outstanding professional color system (cohesive palette with intentional contrast and
  harmony, not random or default colors)
- Outstanding micro-interactions and animations (see animation criteria below)
- Maximum possible score: 8.0

STEP 3: Check for Defects
Evaluate the page for these common problems and factor them into the score:
- Large areas of high-saturation or clashing colors causing visual fatigue
- Responsive/layout issues visible in screenshot (content misaligned, huge blank areas,
  cramped sections)
- Text with insufficient contrast ratio making it difficult to read
- Misalignment issues (buttons in a row at different vertical positions, inconsistent card
  spacing)
- Overlapping or overflowing elements (text spilling outside boundaries, images covering
  text)
- Misused blank space (huge empty areas while content is cramped elsewhere, navigation
  taking excessive screen height)
- Broken images, placeholder images, or pixelated/blurry icons -- worse if it's a main hero
  image
- Buttons/cards lacking modern visual affordances (completely flat with no border-radius,
  shadows, or hover states)

STEP 4: Animation & Motion Quality Assessment
- 7.0+ requires purposeful, sophisticated animations -- not just elements bouncing, sliding
  in, or fading. The animations must serve a clear UX purpose: guiding user attention,
  providing interaction feedback, indicating state changes, or improving navigation flow.
- Penalize overdesign: Gratuitous animations (random bouncing elements, distracting
  continuous motion with no purpose, excessive parallax, elements animating just because
  they can) should LOWER the score, not raise it. Motion without purpose is a defect, not a
  feature.
- Simple CSS transitions on hover/focus are baseline expectations, not bonus-worthy.

STEP 5: Apply Reality Check
6.0-6.5 Score Examples (NOT 8.0):
- Uses Ant Design/Material-UI/Bootstrap with default configuration
- Layout is neat but lacks unique visual identity
- Safe but bland color scheme (standard blue-white-gray)
- Everything is "normal, modern" but nothing makes you say "wow"

7.0-7.5 Score Examples (Close but NOT 8.0):
- Has custom design but 1-2 small flaws
- Overall excellent but one section feels cramped or sparse
- Advanced color scheme but icon styles are inconsistent
- Professional but missing that "extra polish"
- Has good animations that are purposeful but not exceptional

8.0 Score Must Have ALL of these:
- Every element's position feels "exactly right"
- Deliberate color scheme with clear design intent
- At least 3 "beyond standard" design details
- All animations serve clear UX purposes and feel polished, not gratuitous
- Reaches "if this were a SaaS product landing page, I'd believe it's worth paying for"
  quality
- Reference level: Apple.com, Stripe Dashboard, Linear App, Vercel homepage

Final Aesthetics Score: Synthesize baseline, bonus qualities, defects, and animation
assessment into a single score on the 0.0-8.0 scale (capped at 8.0). Explain your reasoning
step by step with a clear calculation breakdown.

III. FINAL OUTPUT FORMAT
Respond ONLY with a JSON object. The reason fields MUST include:
1. Detailed step-by-step explanation
2. Clear deduction breakdown with math

{
    "functional_reason": "Step 1: [Base score, reason]. Step 2: [Instruction following audit
        - query alignment, elaboration, hallucination, template check, language]. Step 3:
        [Source code interactive elements verification]. Step 4: [Data display check]. Step 5:
        [Console error audit]. Calculation: [show math]. Final: X.X",
    "functional_score": 0.0,
    "aesthetics_reason": "Step 1: [Baseline score, reason]. Step 2: [Bonus qualities]. Step 3:
        [Defects found]. Step 4: [Animation assessment]. Step 5: [Reality Check]. Calculation:
        [show math]. Final: Y.Y",
    "aesthetics_score": 0.0
}
\end{Verbatim}
\end{tcolorbox}

\subsubsection{Stage 3 --- Video-Based Problem Detection Prompt}
\label{app:judge_prompts:video_detect}

\begin{tcolorbox}[
    enhanced,
    breakable,
    colback=white,
    colframe=black,
    title=\textbf{Prompt: Video-Based Problem Detection (Stage 3a)},
    fonttitle=\small,
]
\begin{Verbatim}[fontsize=\scriptsize, breaklines=true, breakanywhere=true]
Video-Based Problem Detection

You are an expert web application evaluator. Your task is to analyze the agent's interaction
video and identify any problems or issues that were not visible in the initial static
screenshot.

Context

User Requirements:
{game_description}

Initial Static Evaluation:
- Aesthetics Score: {static_aesthetics_score}/8
- Aesthetics Reason: {static_aesthetics_reason}
- Functional Score: {static_functional_score}/8
- Functional Reason: {static_functional_reason}

Agent Interaction Log (browser_actions):
The following is the sequence of actions the agent performed on the application. Each entry
contains:
- step: sequential step number
- tool: the browser tool used (e.g. browser_navigate, browser_vision_control)
- arguments: parameters passed to the tool (e.g. URL navigated to, thought process and
  action for vision control)

{interaction_summary}

IMPORTANT: Scope of Video Testing

The video interaction is LIMITED in scope:
- For real-time interactive games (e.g., Snake, Flappy Bird, racing games): The agent only
  tests the most basic functional logic (e.g. start, basic movement). Advanced features,
  edge cases, and complex interactions are NOT tested.
- For turn-based or form-based applications: The agent tests more comprehensively but still
  may miss edge cases.
- The video does NOT exhaustively test all features - it's a sampling of typical user
  interactions.

Therefore, when evaluating:
1. Confirm problems found in video - these are DEFINITE issues
2. Do NOT dismiss code-level problems from static evaluation - unless the video explicitly
   proves they don't exist
3. If static evaluation identified a problem in the code, assume it exists UNLESS:
   - The video explicitly demonstrates the feature working correctly
   - The video shows the problematic code path executing without issues

Your Task
Carefully review the initial screenshot, the source code context, and the interaction video
/frames. Identify:
1. NEW problems discovered in video - issues that became apparent during interaction
2. CONFIRMED code problems - issues mentioned in static evaluation that are NOT disproven by
   the video

Focus Areas

Functional Problems

CRITICAL Severity (suggest -2.0 or more each):
- Core logic complete failure: Application crashes, infinite loop, or becomes completely
  unusable (blocks ALL usage)
- Language mismatch: Page language doesn't match user query requirement
- Critical console ERROR with observable malfunction: Runtime error that breaks
  functionality or core logic
  - NOT penalized: External resource loading failures (fonts.googleapis.com,
    fonts.gstatic.com, favicon.ico) with graceful fallback

MAJOR Severity (suggest -1.0 each):
- Non-functional or missing core element: Interactive elements (e.g. buttons, forms) that
  doesn't work at all or missing. If it is very important button (e.g. "buy" for an
  e-commerce page), it's CRITICAL.
- Data completely unreadable: Element overlapping making data invisible
- Key feature missing/broken: Core requirement not working
- Game instruction missing: If the game is not common sense but does NOT have playing
  instructions

MINOR Severity (suggest -0.5 each):
- Secondary button non-functional or with placeholder logic: Help/info buttons or footer
  links. Header navigation bars are NOT secondary.
- Minor readability issues: Small font, low contrast for non-critical text
- Minor implementation gaps: Incomplete form validation, edge case handling

NOT Penalized (DO NOT report as problems):
- Untested cases: Video didn't test it, so DON'T assume it's broken
- Console warnings: Acceptable, not penalized
- External resource failures with graceful fallback: Fonts, favicon loading failures
- Backend/Database operations: Login systems, authentication, data persistence
- Third-party APIs: LLM APIs, payment gateways, social media APIs, map services
- File operations: File upload/download functionality
- Email/SMS: Email sending, SMS verification

Aesthetic Problems

CRITICAL Severity (suggest -2.0 or more each):
- Large high-saturation color area: Neon purple/cyan/green covering large area causing
  visual fatigue
- Responsive issues: Content left-aligned leaving huge blank area on right half of screen

MAJOR Severity (suggest -1.0 each):
- Insufficient text contrast: ANY text difficult to read due to poor contrast ratio
- Misalignment: Buttons in row at significantly different vertical heights, cards with
  inconsistent spacing
- Overlapping/overflowing elements: Text spilling outside boundaries, images covering text,
  NOT intentional design
- Misused blank space: Huge empty area in center while content cramped in corners; excessive
  whitespace breaking flow
- Broken main image: Hero image or critical visual showing broken icon or placeholder

MINOR Severity (suggest -0.5 each):
- Broken secondary image: Decorative or secondary image showing broken icon
- Lack of visual affordances: Completely flat buttons with no border-radius, shadows, or
  hover states

NOT Penalized (DO NOT report as problems):
- Subjective "could be better" opinions: Not objective defects
- Missing advanced design details: These earn +bonus, not -deduction
- "Not 8.0 level": That's normal; 8.0 requires exceptional quality beyond standard

Critical Rules

1. Match severity to EXACT deduction amounts: Don't guess - use the mappings above
2. Report NEW problems from video: Issues that became visible during interaction
3. Retain untested code problems: If static found issue and video didn't test it, RETAIN it
4. Don't penalize untested scenarios: No edge case testing = no edge case deduction
5. Be objective and specific: Only report observable defects matching the criteria above
6. Check Environment Limitations: Don't penalize backend/API/file operations that can't work
   in test environment

Output Format
Return a JSON object with this structure:

{
  "functional_problems": [
    {
      "severity": "CRITICAL|MAJOR|MINOR",
      "description": "Clear description of the problem",
      "timestamp": "When it occurred in the interaction",
      "note": "Optional: Explanation if retained from static evaluation"
    }
  ],
  "aesthetic_problems": [
    {
      "severity": "CRITICAL|MAJOR|MINOR",
      "description": "Clear description of the problem",
      "timestamp": "When it occurred in the interaction",
      "note": "Optional: Explanation if retained from static evaluation"
    }
  ],
  "dismissed_static_problems": [
    {
      "type": "functional|aesthetic",
      "original_issue": "What the static evaluation claimed",
      "reason": "Why the video proves this is incorrect"
    }
  ],
  "overall_assessment": "Summary considering both video findings and static verifier issues"
}
\end{Verbatim}
\end{tcolorbox}

\subsubsection{Stage 3 --- Score Adjustment Prompt}
\label{app:judge_prompts:video_adjust}

\begin{tcolorbox}[
    enhanced,
    breakable,
    colback=white,
    colframe=black,
    title=\textbf{Prompt: Score Adjustment Based on Video Problem Detection (Stage 3b)},
    fonttitle=\small,
]
\begin{Verbatim}[fontsize=\scriptsize, breaklines=true, breakanywhere=true]
Score Adjustment Based on Video Problem Detection

You are an expert evaluator tasked with adjusting initial static evaluation scores based on
problems discovered during interactive testing video.

Context

User Requirements:
{game_description}

Initial Static Evaluation:
- Aesthetics Score: {static_aesthetics_score}/8
- Aesthetics Reason: {static_aesthetics_reason}
- Functional Score: {static_functional_score}/8
- Functional Reason: {static_functional_reason}

Problems Detected in Video:
{detected_problems}

Your Task
Based on the problems discovered during video interaction, adjust the static scores
following the ORIGINAL SCORING STANDARDS below, then provide a UNIFIED final reason that
combines both the original static assessment and the adjustments.

ORIGINAL SCORING STANDARDS

FUNCTIONAL SCORING (0.0 - 8.0 SCALE)

Base Score Tiers:
- 8.0: ALL user requirements fully implemented with correct logic
- 7.0: ALL requirements implemented, only minor features missing/error
- 5.0: Core requirements implemented, some key features missing/error
- 3.0: Core logic incomplete
- 1.0: Only skeleton code, all interactions missing/erroneous
- 0.0: Page fails to render, is blank, shows raw code, only placeholder page, or completely
  irrelevant to query

Interactive Elements Deductions:
- -1.0 per button non-functional or with placeholder logic (alert/console.log only); if it
  is very critical button, deduct MORE.
- -0.5 per secondary button non-functional or with placeholder logic

Data Display Deductions:
- -0.5: Minor readability issues
- -1.0: Element overlapping, text overflowing or data completely unreadable

Mandatory Deductions:
- -2.0: Base deduction using English in the page for queries with non-English languages.
  A WORST case is using non-English language that is not required, punished -4.0.
- -1.0 per CRITICAL console ERROR causing observable malfunction

AESTHETIC SCORING (0.0 - 8.0 SCALE)

Base Score Tiers:
- 5.0: High-quality, clean, modern (standard professional)
- 3.0: Functional but unpolished
- 1.0: Raw HTML with no/minimal styling
- 0.0: Page fails to render, is blank, shows raw code, only placeholder page, or completely
  irrelevant to query

Bonus Points (up to +3.0):
- +1.0: VERY outstanding visual hierarchy
- +1.0: VERY outstanding professional color system
- +1.0: VERY outstanding micro-interactions

Mandatory Deductions:
- -2.0: Large area using high-saturation colors causing visual fatigue
- -2.0: Responsive issues (e.g., content aligned left with huge blank area)
- -1.0: ANY text with insufficient contrast
- -1.0 EACH: Misalignment issues
- -1.0 EACH: Overlapping/overflowing elements (not intentional)
- -1.0 EACH: Misused blank space
- -0.5 to -1.0 EACH: Broken/placeholder/pixelated images
- -0.5: Buttons/cards lack modern visual affordances

VIDEO PROBLEM MAPPING TO STANDARD DEDUCTIONS

Programming language specification: The output project MUST be in html/react WHATEVER the
query specified. DO NOT DEDUCT points for the difference from user query.
Page language check: The web page's displayed language MUST follow the query requirements;
if not specified, it should be the same as the user query.

Functional Problems -> Deductions (from Original Standards)

CRITICAL Severity (>= -2.0 per issue):
- Complete application failure: App crashes on load, infinite loop blocking ALL usage ->
  -4.0 or more (may drop to 0.0)
- Core feature completely broken: Game unplayable, core logic failed -> -3.0 (drops from
  8.0/7.0 -> ~3.0-4.0 tier)
- Language mismatch: Page language doesn't match user query requirement -> -2.0 (mandatory)

MAJOR Severity (-1.0 per issue):
- Critical button with placeholder logic: Submit/login/checkout/save has only alert() or
  console.log() -> -1.0
- Non-functional or missing required element: Interactive element doesn't work -> -1.0
- Data completely unreadable: Element overlapping making data invisible -> -1.0
- Key feature broken: Important requirement not working -> -1.0 (contributes to tier drop)
- Critical console ERROR with observable malfunction: Error that breaks functionality or
  core logic -> -1.0 EACH (Exclude external resource loading failures)

MINOR Severity (-0.5 per issue):
- Secondary button with placeholder logic: Help/info/minor buttons only have alert/console
  .log -> -0.5
- Minor readability issues: Small font, low contrast for non-critical text -> -0.5
- Minor implementation gaps: Incomplete form validation (not tested with invalid input) ->
  -0.5 (drops from 8.0 -> 7.0)

NOT Deductible (DO NOT apply):
- Untested edge cases, Console warnings, Backend/API/file operations, Subjective "could be
  better".

Aesthetic Problems -> Deductions (from Original Standards)

CRITICAL Severity (-2.0 per issue):
- Large high-saturation color area: Neon purple/cyan/green causing visual fatigue -> -2.0
- Responsive layout failure: Content left-aligned leaving huge blank on right -> -2.0

MAJOR Severity (-1.0 per issue):
- Insufficient text contrast: ANY text difficult to read -> -1.0
- Misalignment: Buttons at different heights, inconsistent card spacing -> -1.0 EACH
- Overlapping/overflowing elements: Text spilling outside, unintentional overlap -> -1.0
  EACH
- Misused blank space: Huge empty area in center while content cramped -> -1.0 EACH
- Broken main image: Hero image or critical visual broken -> -1.0

MINOR Severity (-0.5 per issue):
- Broken secondary image: Decorative image broken -> -0.5
- Lack of visual affordances: Completely flat buttons with no styling -> -0.5

NOT Deductible (DO NOT apply):
- Subjective opinions, Missing advanced design details, "Not 8.0 quality".

ADJUSTMENT PRINCIPLES

1. No Double-Penalization (CRITICAL RULE)
If static evaluation already deducted for an issue, DO NOT deduct again:
- Static says "-1.0 for overlapping text" + video confirms it -> No additional deduction
- Static says "-0.5 for incomplete validation" + video confirms it -> No additional
  deduction
- Only deduct for NEW problems discovered in video that weren't mentioned in static

2. Untested != Broken
If video didn't test something, DON'T assume it's broken:
- Video shows valid form submission -> DON'T deduct for "missing validation" (wasn't tested)
- Video shows 30s gameplay -> DON'T deduct for "missing win condition" (wasn't reached)
- Exception: If static found it in code and video didn't contradict, RETAIN the static
  deduction

3. Context Matters
- Static LOW (<=3.0) + video problems -> Small additional deduction
- Static HIGH (>=6.0) + video reveals critical crash -> Large deduction warranted

4. Score Floor
Minimum adjusted score is 0.0 (don't go negative). If page renders correctly and is
relevant to the query, it should be at least 1.0 for both functional and aesthetics.

5. Positive Adjustments
Only if video PROVES static was wrong:
- e.g. static said "button broken" but video shows it works perfectly -> Can add back
  deducted points
- Must justify clearly

6. Deduction Restrictions
- Do NOT let the final score significantly lower than its query requirement completeness
- e.g. A fully implemented web page but all buttons non-functional should NOT be 1.0 for
  functional, suggested 3.0-4.0
- NOT suitable for game loop elements.

Your Task
Based on the static evaluation scores and the problems detected in the video, calculate
appropriate adjusted scores following the ORIGINAL SCORING STANDARDS, and provide UNIFIED
reasons that combine both the original assessment and the adjustments.

Remember:
1. Follow the original scoring scale and deduction rules
2. Don't double-penalize issues already in static evaluation
3. Provide unified reasons that are self-contained and clear
4. Show clear calculation math in the reasons
5. Be thorough, fair, and consistent with the guidelines

OUTPUT FORMAT

Return a JSON object with this structure:

{
  "adjusted_functional_score": 5.5,
  "functional_reason": "UNIFIED reason combining original assessment and adjustments. Format:
    [Original static assessment summary]. [Video interaction findings]. [Final calculation
    with clear deduction math]. Final Score: X.X/8.0",
  "adjusted_aesthetics_score": 6.0,
  "aesthetics_reason": "UNIFIED reason combining original assessment and adjustments. Format:
    [Original static assessment summary]. [Video interaction findings]. [Final calculation
    with clear deduction math]. Final Score: Y.Y/8.0",
  "adjustment_summary": "Brief 1-2 sentence summary of overall adjustment decision"
}
\end{Verbatim}
\end{tcolorbox}


\subsection{Human Annotation Rubric}
\label{app:human_rubric}

This appendix documents the 16-item rubric used by our human annotators and the principles behind it. As motivated in Section~\ref{sec:eval_dims}, the rubric is not a target that the machine judge is asked to reproduce item by item. It serves as a fine-grained calibration instrument whose aggregated scores along \emph{functionality} and \emph{aesthetics} are used in Section~\ref{sec:ablation} to quantify machine-human agreement.

\subsubsection{Design Principles}
\label{app:human_rubric:principles}

Three principles shape the rubric. First, each item is \textbf{rendered as a binary pass or fail}: human judgment is asked to return 0 or 1 per item, not a graded opinion. Continuous scoring is deferred to the aggregated dimension level, where it is produced arithmetically rather than cognitively. Second, each item admits \textbf{scene-adapted criteria}: the generic web criterion is refined per scene, because the same visual behaviour can be correct in one scene and defective in another. Fourteen scenes are covered: generic web front-end, game, clone, tool, landing page, creative, e-commerce, blog, map, dashboard, data visualisation, 3D and animation, UI components, and SVG-driven page. Third, each item carries \textbf{explicit exemption rules} that neutralise failures attributable to the execution sandbox rather than to the model. The main exemption families are missing a server or authentication backend, absent third-party APIs including payment and external services, and forms that cannot submit against a real endpoint. Functionality expressible with React state, local storage, or mock data is \emph{not} exempt, so that front-end logic remains fully evaluated.

\subsubsection{Item Inventory}
\label{app:human_rubric:items}

Table~\ref{tab:human_rubric_items} lists the 16 items grouped by their original thematic cluster, the dimension each contributes to at the aggregation stage, and a one-sentence pass criterion. Full scene-adapted criteria and exemption rules are omitted here for space and will be released together with the annotation interface.

\begin{table*}
\centering
\footnotesize
\caption{Sixteen-item human annotation rubric. Suffixes 2a and 3a are refinement sub-items for language consistency and feature-count auditing, scored separately at aggregation. All items are binary (0/1) per task.}
\label{tab:human_rubric_items}
\renewcommand{\arraystretch}{1.15}
\setlength{\tabcolsep}{3pt}
\begin{tabular}{c p{2.6cm} p{2.4cm} c p{7.0cm}}
\toprule
\textbf{\#} & \textbf{Cluster} & \textbf{Item} & \textbf{Dim.} & \textbf{Pass Criterion} \\
\midrule
1  & Functionality         & Code renders                & F & Page loads and React or scaffold project builds without fatal errors. \\
2  & Functionality         & Intent alignment            & F & Generated page type, content, visual style, and media match the user query. \\
2a & Functionality         & Language consistency        & F & Interface copy, prompts, buttons, and error messages match the query language. \\
3  & Functionality         & Logic correctness           & F & Interactive widgets, routing, state, and scene-specific workflows behave as specified. \\
3a & Functionality         & Feature count               & F & Numeric tally of correctly and incorrectly realised features for auditing. \\
4  & Functionality         & Data display                & F & No truncation, overflow, overlap, or character-set corruption in rendered content. \\
5  & Functionality         & No console errors           & F & Browser console reports no runtime errors; warnings are exempt. \\
6  & Functionality         & Responsive adaptation       & F & Layout remains intact across mobile (375px), tablet (768px), and desktop (1280px) breakpoints. \\
\midrule
7  & Aesthetics (static)   & Layout rationality          & A & Information density is balanced; hierarchy and module separation are clear. \\
8  & Aesthetics (static)   & Interface regularity        & A & Typography, spacing, alignment, and component sizes follow a consistent grid. \\
9  & Aesthetics (static)   & Colour harmony              & A & Saturation, contrast, and palette unity serve the scene rather than fragment it. \\
10 & Aesthetics (static)   & Design refinement           & A & Page exhibits design intent beyond bare content display, with polished detailing. \\
\midrule
11 & Interactivity         & Animation smoothness        & A & State transitions and game-loop frames run without jank; load latencies are reasonable. \\
12 & Interactivity         & Transition effects          & A & Hover, expand, modal, and menu transitions use coherent easing rather than hard switches. \\
13 & Interactivity         & Interaction feedback        & F & Every user action elicits immediate visible feedback and actionable error messages. \\
14 & Interactivity         & User experience             & F & Controls are intuitive, latencies remain under a perceptible threshold, and flows terminate cleanly. \\
\midrule
15 & Content quality       & Image asset quality         & A & Images load, resolutions match the slot, and imagery is consistent with surrounding copy. \\
16 & Content quality       & Audio and video behaviour   & F & Media assets load, trigger on the right events, expose volume control, and do not conflict. \\
\bottomrule
\end{tabular}
\end{table*}

\subsection{Aggregation to Two Dimensions}
\label{app:human_rubric:aggregation}

Let $x_j \in \{0, 1\}$ be the binary score of item $j$ after applying the scene-adapted criterion and any active exemption. The two dimension scores per task are produced as uniform averages over the relevant item subsets:
\begin{equation}
s^{\mathrm{F}}_i = \frac{1}{|\mathcal{I}_{\mathrm{F}}|} \sum_{j \in \mathcal{I}_{\mathrm{F}}} x_j^{(i)},
\qquad
s^{\mathrm{A}}_i = \frac{1}{|\mathcal{I}_{\mathrm{A}}|} \sum_{j \in \mathcal{I}_{\mathrm{A}}} x_j^{(i)},
\end{equation}
where $\mathcal{I}_{\mathrm{F}}$ and $\mathcal{I}_{\mathrm{A}}$ denote the item subsets recorded in the \textbf{Dim.} column of Table~\ref{tab:human_rubric_items}. All applicable items within a dimension contribute equally; items rendered inapplicable by an active exemption are dropped from both the sum and the count rather than being scored as failures, preventing exempt features from diluting the dimension.

\section{Worked Example: \dataname Evaluation Trace}
\label{app:case_study}

This appendix walks through a complete evaluation trace produced by \framename, showing every artifact the verifier receives and every scoring decision it makes. The example is a single query (``Super Mario'') executed by Claude-Opus-4.7 in HTML mode. The trace is representative: it illustrates how static perception, agent-driven interaction, and deferred scoring combine to surface defects that would be invisible to either stage in isolation.

\subsection{Query and Generated Output}
\label{app:case_query}

\paragraph{User query.} \texttt{Super Mario}

\paragraph{Model output.} A single-file HTML/CSS/JS browser-based platformer game implementing a complete ``title~$\rightarrow$~play~$\rightarrow$~win/lose~$\rightarrow$~restart'' loop. The game includes gravity physics, collision detection, jumping, enemy stomping (Goomba), power-ups (mushrooms), coin collection from question blocks, a parallax background, particle effects, and a Web Audio API sound system. Keyboard controls (WASD / Arrow keys / Space / Enter) and mobile touch controls are both implemented.

\subsection{Stage~1: Static Perception}
\label{app:case_static}

\subsubsection{Inputs to the Verifier}

\begin{figure}[h]
\centering
\includegraphics[width=0.85\linewidth]{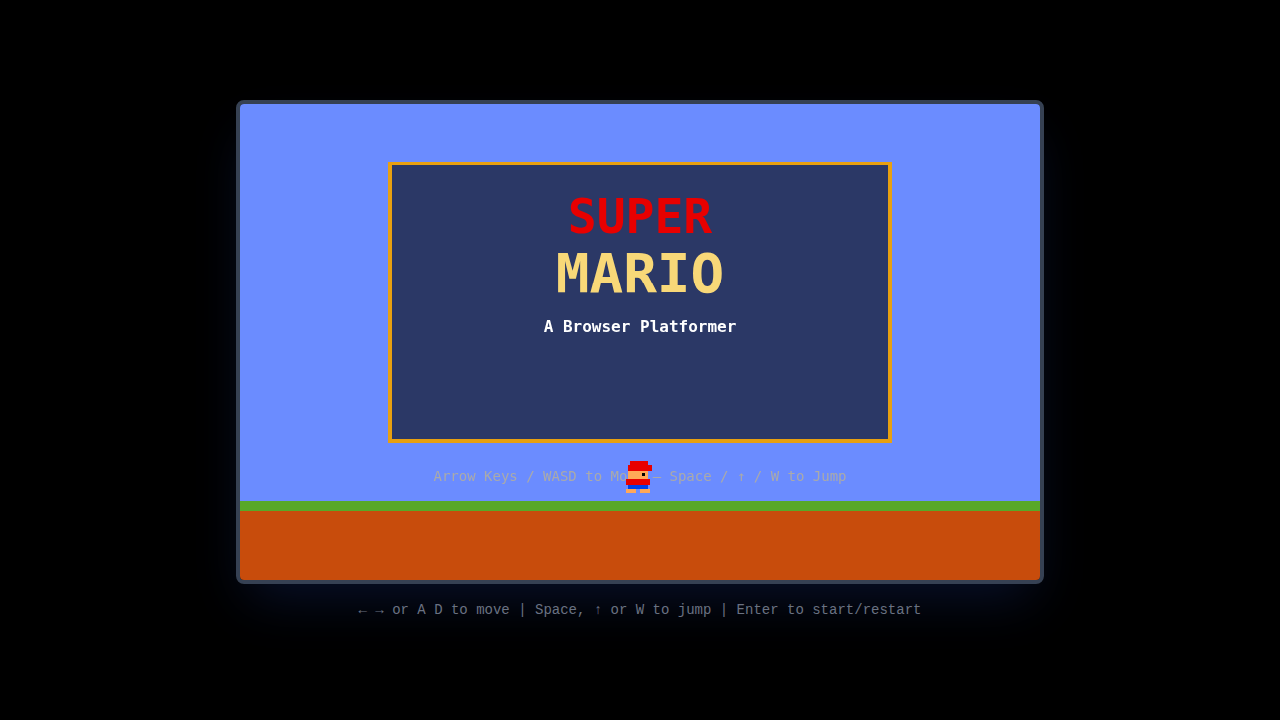}
\caption{Rendered screenshot of the generated Super Mario game as seen by the static verifier. The title screen shows parallax clouds, hills, and bushes; the HUD displays score, coins, time, and lives; control instructions appear at the bottom.}
\label{fig:mario-static}
\end{figure}

At Stage~1 the verifier receives four inputs: (1)~a rendered screenshot of the landing page (Figure~\ref{fig:mario-static}), (2)~the complete source code of the single HTML file, (3)~the original user query, and (4)~the browser console logs (no errors or warnings).

\subsubsection{Static Scoring Output}

The verifier assigns provisional scores on a 0--8 scale.

\begin{table}[h]
\centering
\small
\caption{Stage~1 static scores for the Super Mario example.}
\label{tab:mario-static}
\begin{tabular}{lp{10cm}c}
\toprule
Dimension & Verifier reasoning (abridged) & Score \\
\midrule
Aesthetics & Step~1: Baseline 5.0 for a clean canvas-based game. Step~2: Bonus +2.2 for custom pixel-art renderer (no external assets), parallax background system, Web Audio API sound system, and particle effects. Step~3: Deduction $-$0.2 for instruction text slightly overlapping the Mario sprite on the title screen. Step~4: Animations are purposeful (walking/jumping frames, block bumping, breaking-brick particles). Step~5: High visual consistency despite simplified character design. \textit{Calculation:} $5.0 + 2.2 - 0.2 = 7.0$. & 7.0/8.0 \\
\addlinespace
Functionality & Step~1: Complete platformer with full game loop, core mechanics (gravity, collision, jumping, enemies, power-ups, blocks), HUD, and controls. Step~2: Output is a playable game, not a landing page, matching the query. Step~3: All interactive elements (keyboard and mobile touch controls) are correctly implemented in source. Step~4: HUD updates real-time game data. Step~5: No console errors. \textit{Calculation:} 8.0 base for complete, bug-free implementation. & 8.0/8.0 \\
\bottomrule
\end{tabular}
\end{table}

\subsection{Stage~2: Agent-Driven Interaction}
\label{app:case_interact}

\subsubsection{Interaction Trajectory}

The computer-using agent receives the same four inputs as the Stage~1 verifier plus the static scores. It then executes a 27-step trajectory to exercise the application. The complete action log is reproduced below; each entry contains the tool call, arguments (including the agent's own reasoning), and timing parameters.

\begin{tcolorbox}[
    enhanced,
    breakable,
    colback=white,
    colframe=black!70,
    title=\textbf{Full Interaction Trajectory (27 steps)},
    fonttitle=\small,
]
\begin{Verbatim}[fontsize=\scriptsize, breaklines=true, breakanywhere=true]
Step 1: browser_navigate
  Arguments: {"url": "http://localhost:44608"}

Step 2: browser_press_key
  Arguments: {"key": "Enter",
    "step": "Press Enter to start the Super Mario game as instructed on the
      start screen",
    "thought": "I see the game's start screen that says 'PRESS ENTER OR SPACE
      TO START', so I need to press Enter to begin the game and test its
      core functionality."}

Step 3: browser_press_key
  Arguments: {"key": "ArrowRight",
    "step": "Press Right Arrow key to move Mario to the right as per the game
      controls",
    "thought": "The game has started, and Mario is on the starting platform.
      According to the controls at the bottom, arrow keys can be used to move,
      so I'll press right arrow to test movement functionality.",
    "duration_ms": 300}

Step 4: browser_press_key
  Arguments: {"key": "ArrowRight",
    "step": "Press Right Arrow key again to continue moving Mario to the right",
    "thought": "Mario moved right after the first right arrow press, so I'll
      continue moving right to test further movement and encounter the question
      blocks.",
    "duration_ms": 300}

Step 5: browser_press_key
  Arguments: {"key": "Space",
    "step": "Press Space to test jumping functionality as per the game controls",
    "thought": "Now I've moved Mario forward, and I need to test jumping. The
      controls say Space can be used to jump, so I'll press Space to see if
      Mario jumps.",
    "duration_ms": 100}

Step 6: browser_press_key
  Arguments: {"key": "ArrowRight",
    "step": "Press Right Arrow key to move Mario towards the question blocks",
    "thought": "Mario successfully jumped when I pressed Space, so now I'll
      continue moving right to reach the question blocks and test interacting
      with them.",
    "duration_ms": 400}

Step 7: browser_press_key
  Arguments: {"key": "ArrowRight",
    "step": "Press Right Arrow key to move Mario closer to the enemy Goomba",
    "thought": "Mario is approaching the Goomba enemy, I'll continue moving right
      to see if Mario can interact with it (defeat it by jumping or collide
      and lose).",
    "duration_ms": 300}

Step 8: browser_press_key
  Arguments: {"key": "Space",
    "step": "Press Space to jump over the Goomba enemy",
    "thought": "Goomba is now on the platform with Mario, I need to jump over it,
      so I'll press Space to jump.",
    "duration_ms": 100}

Step 9: browser_press_key
  Arguments: {"key": "ArrowRight",
    "step": "Press Right Arrow key to move Mario past where Goomba was after
      jumping",
    "thought": "Mario successfully jumped over Goomba, which fell into the pit.
      Now I'll continue moving right to reach the question blocks and hit them.",
    "duration_ms": 400}

Step 10: browser_press_key
  Arguments: {"key": "ArrowRight",
    "step": "Press Right Arrow key to continue moving Mario towards the isolated
      question block",
    "thought": "Mario is moving towards the isolated ? block. I'll keep moving
      right to get into position to jump and hit the block.",
    "duration_ms": 400}

Step 11: browser_press_key
  Arguments: {"key": "ArrowRight",
    "step": "Press Right Arrow key to move Mario under the isolated question
      block",
    "thought": "Mario is getting close to the isolated ? block. I need to position
      him under it so I can jump and hit the block to get a coin.",
    "duration_ms": 400}

Step 12: browser_press_key
  Arguments: {"key": "ArrowRight",
    "step": "Press Right Arrow key to get Mario into position under the question
      block",
    "thought": "I need to get Mario directly under the ? block so when I jump,
      I can hit it. I'll keep moving right to position correctly.",
    "duration_ms": 300}

Step 13: browser_press_key
  Arguments: {"key": "ArrowRight",
    "step": "Press Right Arrow key to move Mario directly under the question
      block",
    "thought": "Mario is almost in the right position. I need to move one more
      step right so I can hit the ? block when I jump.",
    "duration_ms": 200}

Step 14: browser_press_key
  Arguments: {"key": "ArrowRight",
    "step": "Press Right Arrow key one more time to position Mario directly under
      the question block",
    "thought": "Mario needs just a little more movement right to be under the
      ? block. I'll press right arrow once more to get the correct position.",
    "duration_ms": 250}

Step 15: browser_press_key
  Arguments: {"key": "ArrowRight",
    "step": "Press Right Arrow one final time to get Mario under the question
      block",
    "thought": "The previous presses didn't move Mario enough. Let me press
      Right Arrow once more to position him correctly so that when I jump,
      I'll hit the question block.",
    "duration_ms": 300}

Step 16: browser_press_key
  Arguments: {"key": "ArrowRight",
    "step": "Keep moving Mario right towards the question block, he's still not
      close enough",
    "thought": "Mario has moved right but still has a way to go before reaching
      the question block. I need to keep pressing the Right Arrow to get him
      into position.",
    "duration_ms": 500}

Step 17: browser_press_key
  Arguments: {"key": "ArrowRight",
    "step": "Continue moving Mario right towards the question block",
    "thought": "Mario is moving but still needs to get closer to the question
      block. I'll press Right Arrow again to keep moving towards it.",
    "duration_ms": 500}

Step 18: browser_press_key
  Arguments: {"key": "ArrowRight",
    "step": "Press Right Arrow to keep Mario moving towards the question block",
    "thought": "Mario is now over the small bush, but still needs to reach the
      question block. I will continue moving right to get closer.",
    "duration_ms": 500}

Step 19: browser_press_key
  Arguments: {"key": "ArrowRight",
    "step": "Press Right Arrow to move past the bush and get closer to the
      question block",
    "thought": "Mario is now on the small bush, but the bush is part of the
      ground, so he can still move over it. I'll continue pressing right to get
      to the question block.",
    "duration_ms": 500}

Step 20: browser_press_key
  Arguments: {"key": "ArrowRight",
    "step": "Press Right Arrow to continue moving towards the question block",
    "thought": "Mario is still moving right, getting closer to the isolated
      question block. I need to keep pressing right until he's positioned under
      it to jump.",
    "duration_ms": 500}

Step 21: browser_press_key
  Arguments: {"key": "ArrowRight",
    "step": "Press Right Arrow to keep moving towards the question block",
    "thought": "Mario is getting closer to the question block, I just need to
      keep moving right to reach it.",
    "duration_ms": 500}

Step 22: browser_press_key
  Arguments: {"key": "ArrowRight",
    "step": "Continue pressing Right Arrow to move Mario closer to the question
      block",
    "thought": "Mario is almost at the question block, just a little more
      movement to get into the perfect position to jump and hit it.",
    "duration_ms": 500}

Step 23: browser_press_key
  Arguments: {"key": "ArrowRight",
    "step": "Press Right Arrow to keep moving towards the question block",
    "thought": "Mario is now close to the pit before the question block. I need
      to keep moving right until I reach the question block's platform.",
    "duration_ms": 500}

Step 24: browser_press_key
  Arguments: {"key": "ArrowRight",
    "step": "Press Right Arrow to move Mario towards the edge of the platform
      before the pit",
    "thought": "Mario is at the edge of the current platform. I need to jump
      across the gap to the next platform where the question block is. First,
      I'll move as far right as possible on the current platform, then jump.",
    "duration_ms": 400}

Step 25: browser_press_key
  Arguments: {"key": "ArrowRight",
    "step": "Press Right Arrow to get Mario to the very edge of the platform,
      preparing to jump across the gap",
    "thought": "Mario is at the edge now. I need to get him to the absolute
      edge to make the jump across the gap easier. Then I'll press space to
      jump.",
    "duration_ms": 200}

Step 26: browser_press_key
  Arguments: {"key": "ArrowRight",
    "step": "Press Right Arrow to get Mario to the very edge of the platform",
    "thought": "Mario is almost at the edge, just one more press of Right Arrow
      to get him there. Then I can jump across the gap to the other platform.",
    "duration_ms": 200}

Step 27: browser_press_key
  Arguments: {"key": "ArrowRight",
    "step": "Press Right Arrow one more time to get Mario to the very edge",
    "thought": "Now Mario is at the edge, I need to get him one more step right
      to maximize the jump distance. Then I'll press space to jump across the
      gap.",
    "duration_ms": 100}
\end{Verbatim}
\end{tcolorbox}

\subsubsection{Multi-Modal Evidence Captured}

\begin{figure}[h]
\centering
\includegraphics[width=0.85\linewidth]{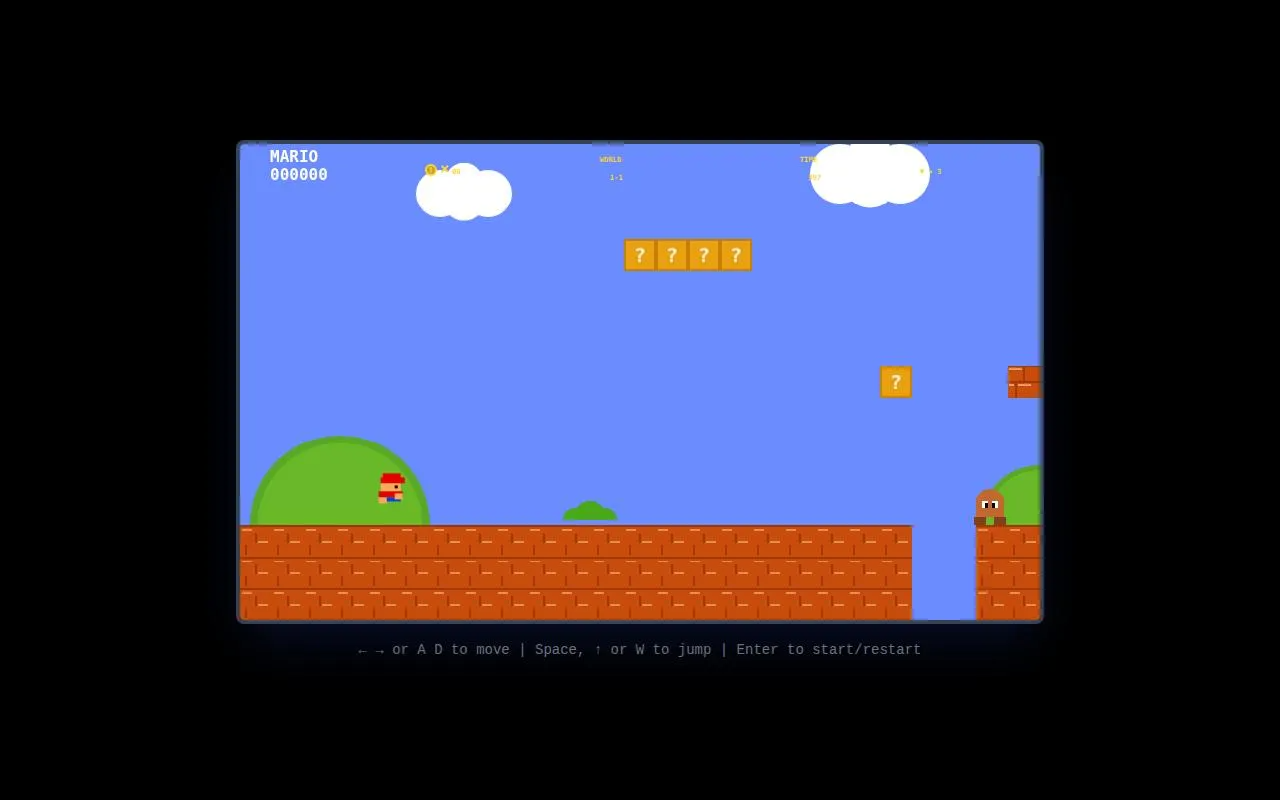}
\caption{Interaction frame captured during the agent's gap-crossing attempt. Mario is positioned at the edge of the first platform; the question block and the gap are visible ahead. The agent repeatedly presses \texttt{ArrowRight} and \texttt{Space} but fails to reach the opposite platform.}
\label{fig:mario-interact}
\end{figure}

During the 27-step interaction the agent records a multi-modal evidence package:
\begin{itemize}
\item A continuous screen-capture video (MP4) of the full gameplay session.
\item Key-frame screenshots at each action boundary (Figure~\ref{fig:mario-interact} shows a representative frame during gap-crossing attempts).
\item The complete action log reproduced above, with per-step tool calls, arguments, duration, and agent reasoning.
\end{itemize}

\subsection{Stage~3: Dynamic Scoring}
\label{app:case_dynamic}

\subsubsection{Problem Detection}

The Stage~3 verifier reasons over the static scores, the interaction video, and the action log. It identifies two problems:

\begin{tcolorbox}[
    enhanced,
    breakable,
    colback=white,
    colframe=black!70,
    title=\textbf{Detected Problems (JSON output)},
    fonttitle=\small,
]
\begin{Verbatim}[fontsize=\scriptsize, breaklines=true, breakanywhere=true]
{
  "functional_problems": [
    {
      "severity": "MAJOR",
      "description": "The physics tuning for jumping is poorly balanced with
        the level design. Mario's horizontal jump distance is insufficient
        to easily clear the first gap, leading to a frustrating user
        experience where the player falls into the pit even when jumping
        from the edge.",
      "timestamp": "00:09",
      "note": "The agent attempted to jump across the first gap multiple
        times but failed to reach the other side, indicating the jump arc
        or movement speed is too low for the provided level layout."
    }
  ],
  "aesthetic_problems": [
    {
      "severity": "MINOR",
      "description": "On the title screen, the instruction text
        ('Arrow Keys / WASD to Move...') overlaps with the Mario character
        sprite.",
      "timestamp": "00:00",
      "note": "Confirmed from static evaluation. This is a minor layout
        defect that reduces the polish of the initial screen."
    }
  ],
  "dismissed_static_problems": [],
  "overall_assessment": "The application is a functional Super Mario clone
    with a complete game loop. However, it suffers from a major functional
    issue regarding physics balance; the jump distance is too short for
    the gaps provided in the level, making progress difficult. A minor
    aesthetic overlap on the title screen is also present."
}
\end{Verbatim}
\end{tcolorbox}

\subsubsection{Score Adjustment}

The verifier applies the adjustment rules (no double-penalization, untested~$\neq$~broken) to produce the final scores.

\paragraph{Aesthetics.} The static evaluation already deducted 0.2 for the title-screen text overlap. Video interaction confirmed this overlap but discovered no \emph{new} aesthetic defects. By the no-double-penalization rule, the score remains unchanged: $5.0 + 2.2 - 0.2 = 7.0$.

\paragraph{Functionality.} The static evaluation awarded 8.0 for a complete, bug-free game. Video interaction revealed a \textbf{MAJOR} functional issue: the jump physics are poorly balanced against the level design, making the first gap nearly impossible to clear---a significant barrier to gameplay. This warrants a $-$1.0 deduction per the scoring rubric. Final: $8.0 - 1.0 = 7.0$.

\begin{table}[h]
\centering
\small
\caption{Final calibrated scores after Stage~3 for the Super Mario example.}
\label{tab:mario-final}
\begin{tabular}{lccc}
\toprule
Dimension & Stage~1 (Static) & Stage~3 (Adjusted) & Change \\
\midrule
Aesthetics  & 7.0 & 7.0 & $-$ \\
Functionality & 8.0 & 7.0 & $-$1.0 (MAJOR: physics/level-design imbalance) \\
\bottomrule
\end{tabular}
\end{table}

\subsection{What the Video Surfaced That Static Inspection Missed}
\label{app:case_video}

The static verifier examined the source code and saw that jumping logic, collision detection, and level geometry were all present. From the code alone, the game appeared fully functional (hence the 8.0). The interaction video, however, revealed an \emph{emergent} defect: the combination of jump arc, horizontal movement speed, and gap width made the first obstacle practically impassable. This is a physics-tuning failure, not a missing feature, and it is only discoverable through embodied interaction---exactly the gap \framename is designed to close.
\section{Detailed Generation Results}
\label{app:generation}

\begin{figure*}[htbp]
\centering
\vspace{-2mm}
\includegraphics[width=\textwidth]{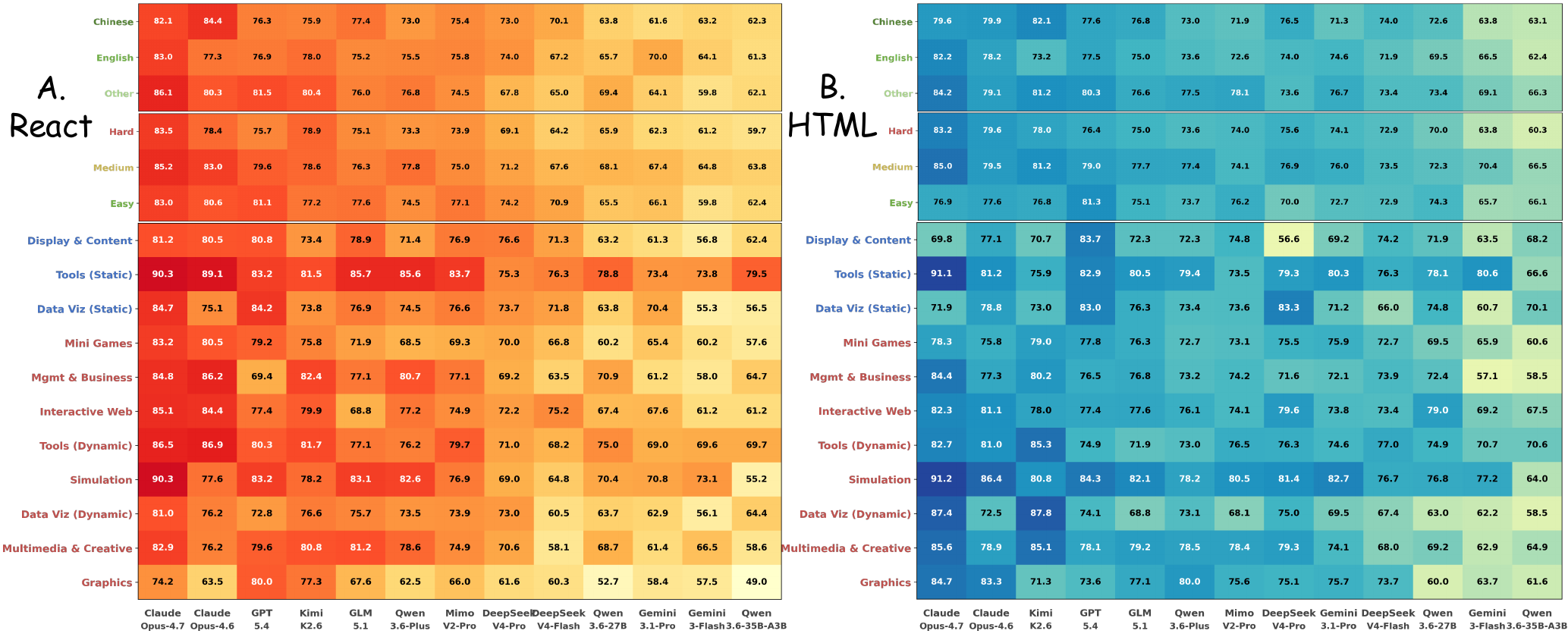}
\caption{Per-model average scores across language, difficulty tier, and L2 category under React (top half) and HTML (bottom half). Darker cells indicate higher scores.}
\label{fig:cross-heatmap}
\vspace{-4mm}
\end{figure*}

Figure~\ref{fig:cross-heatmap} reports per-model average scores across language, difficulty tier, and L2 category under React (top half) and HTML (bottom half). Darker cells indicate higher scores.

\textbf{Difficulty.} An instructive non-monotonicity appears along the difficulty axis: medium tasks score highest on average, with easy and hard tasks trailing on both sides. We attribute this to a task--capability mismatch at the extremes. Hard tasks demand precise multi-step state management and constraint satisfaction that exceeds current models' instruction-following fidelity; incomplete implementations or subtle logical errors dominate the failure mode. Easy tasks, conversely, suffer from under-constraint: a brief query such as ``build a personal homepage'' offers so little specification that models over-engineer, inject unsolicited features, or diverge from unstated user intent, and the verifier penalizes the mismatch. Medium tasks sit in a sweet spot where the query is specific enough to guide generation without requiring reasoning depth beyond the model's reliable horizon.

\textbf{L2 category.} Across L2 categories, Tools~(Static) is the universal strength because deterministic widgets with clear completion criteria align well with scaffold-based generation, while Graphics and Animation remain the universal weakness due to the fine-grained spatial and temporal reasoning they demand.

\textbf{Language.} Language effects are weaker than difficulty effects: in React several models peak on non-English prompts, whereas HTML shows a more even distribution with no clear monolingual advantage, suggesting that scaffold structure rather than prompt language is the dominant variable.


\end{document}